\documentclass[10pt,twocolumn]{article}
\usepackage[super,sort&compress]{natbib}
\usepackage[a4paper, left=1.5cm, right=1.5cm, top=2cm, bottom=2cm, columnsep=0.7cm]{geometry}
\usepackage{graphicx}
\usepackage{algorithm}
\usepackage{algorithmic}
\usepackage{subcaption}
\usepackage{amsmath}
\usepackage{float}
\usepackage{tabularx}
\usepackage{amssymb}
\usepackage{multirow} 
\usepackage{booktabs} 
\usepackage{tcolorbox}  
\usepackage[compact]{titlesec}
\usepackage{url}
\usepackage[font=small]{caption} 
\usepackage{hyperref}

\titleformat{\section}
{\normalfont\fontsize{12}{15}\bfseries}{\thesection}{1em}{}

\titleformat{\subsection}
{\normalfont\fontsize{10}{12}\bfseries}{\thesubsection}{1em}{}

\titleformat{\subsubsection}
{\normalfont\fontsize{10}{12}\itshape}{\thesubsubsection}{1em}{}

\begin{document}
	
	\twocolumn[{
		\centering
		\LARGE \textbf{Interpretable and Granular Video-Based Quantification of Motor Characteristics from the Finger Tapping Test in Parkinson Disease 
		} \\
		\vspace{0.3cm}
		\small Tahereh Zarrat Ehsan\textsuperscript{1*}, Michael Tangermann\textsuperscript{2}, Yağmur Güçlütürk\textsuperscript{2}, Bastiaan R. Bloem\textsuperscript{1}, Luc J. W. Evers\textsuperscript{1}\\ 
		\vspace{0.3cm}
		\normalsize
        \small\textsuperscript{1} Department of Neurology, Center of Expertise for Parkinson and Movement
        Disorders, Donders Institute for Brain, Cognition and Behaviour, Radboud
        University Medical Center, Nijmegen, The Netherlands.  \\
        \textsuperscript{2}
        Department of Artificial Intelligence, Donders Institute for Brain, Cognition and Behaviour, Radboud University, Nijmegen, The Netherlands \\
		\href{mailto:tahereh.zarratehsan@radboudumc.nl}{*tahereh.zarratehsan@radboudumc.nl}

		\begin{tcolorbox}[colback=gray!20, colframe=gray!20, sharp corners, boxrule=0pt, width=\textwidth]
			\textbf{Abstract} \\

			Accurately quantifying motor characteristics in Parkinson disease (PD) is crucial for monitoring disease progression and optimizing treatment strategies. The finger-tapping test is a standard motor assessment. Clinicians visually evaluate a patient’s tapping performance and assign an overall severity score based on tapping amplitude, speed, and irregularity. However, this subjective evaluation is prone to inter- and intra-rater variability, and does not offer insights into individual motor characteristics captured during this test. This paper introduces a granular computer vision-based method for quantifying PD motor characteristics from video recordings. Four sets of clinically relevant features are proposed to characterize \textit{hypokinesia}, \textit{bradykinesia}, \textit{sequence effect}, and \textit{hesitation-halts}. We evaluate our approach on video recordings and clinical evaluations of 74 PD patients from the Personalized Parkinson Project. Principal component analysis with varimax rotation shows that the video-based features corresponded to the four deficits. Additionally,
            video-based analysis has allowed us to identify further granular distinctions within \textit{sequence effect} and \textit{hesitation-halts} deficits. In the following, we have used these features to train machine learning classifiers to estimate the Movement Disorder Society Unified Parkinson Disease Rating Scale (MDS-UPDRS) finger-tapping score. Compared to state-of-the-art approaches, our method achieves a higher accuracy in MDS-UPDRS score prediction, while still providing an interpretable quantification of individual finger-tapping motor characteristics. In summary, the proposed framework provides a practical solution for the objective assessment of PD motor characteristics, that can potentially be applied in both clinical and remote settings. Future work is needed to assess its responsiveness to symptomatic treatment and disease progression.
			
		\end{tcolorbox}
		\vspace{0.5cm}
	}]

	\section*{Introduction}
	Parkinson disease (PD) is the second most common neurodegenerative disorder worldwide and is characterized by a progressive loss of dopamine-producing neurons in the brain~\cite{feigin2019global}. Among its hallmark motor signs, bradykinesia is the most prevalent, affecting approximately 80\%~of individuals with PD~\cite{morris2000movement}. Traditionally, bradykinesia has referred to the slow execution of movement, reduced amplitude, and a progressive decrement in speed and amplitude during repetitive actions. However, recent studies~\cite{bologna2023redefining, bologna2023further} suggest that bradykinesia includes potentially distinct motor characteristics with discrete pathophysiological mechanisms. To reflect this, a new revised definition of bradykinesia has been proposed~\cite{bologna2023redefining} which identifies four key motor characteristics that define the motor phenotype of PD and atypical parkinsonism: hypokinesia, bradykinesia, sequence effect, and hesitation-halts. In this framework, bradykinesia is defined as the slowness of movement execution, while hypokinesia refers to a reduction in movement amplitude. The sequence effect describes a progressive reduction in amplitude and speed during repetitive tasks. Hesitation-halts reflect irregularities in movement rhythm, including pauses, hesitations, and fluctuations in timing and amplitude. Bologna et al.~\cite{bologna2023redefining} emphasized that these four mentioned PD deficits do not share a common pathophysiological background as they respond differently to dopaminergic drugs or deep brain stimulation (DBS). Also, Giulia et al.~\cite{paparella2023may} argued that specific combinations of these deficits may aid differential diagnosis. For example, the presence of bradykinesia alongside a sequence effect is highly indicative of Parkinsonism, whereas isolated bradykinesia may occur in a range of neurological conditions~\cite{paparella2023may}. Notably, the sequence effect typically is absent in disorders such as essential tremor~\cite{bologna2020there, passaretti2024role} and progressive supranuclear palsy (PSP)~\cite{ling2012hypokinesia}. As a result, recent literature~\cite{laurencin2024noradrenergic} recommends that these motor deficits be reported separately to improve the precision of disease monitoring and facilitate the development of targeted treatment strategies.
	
	The finger-tapping test is a widely used clinical tool for assessing PD motor characteristics~\cite{goetz2008movement}. During the test, patients are instructed to tap their thumb and index finger as widely and rapidly as possible for a period of 10 seconds, while clinicians visually evaluate their performance based on specific criteria outlined in the Movement Disorder Society Unified Parkinson's Disease Rating Scale (MDS-UPDRS, see table~\ref{table:mds_MDS-UPDRS}). The finger-tapping test results in a single score that reflects the severity of motor impairment. However, clinical ratings remain subjective and prone to inter-rater variability, as different clinicians may interpret subtle movement abnormalities differently~\cite{richards1994interrater, berlot2021variability}. Moreover, the final score summarizes overall performance and does not distinguish between different motor characteristics that may be present during the task. In practice, clinicians may find it challenging to identify and differentiate all relevant motor deficits by visual inspection alone~\cite{angelini2024distinguishing, guerra2023objective}. These limitations highlight the need for an objective and automated assessment method capable of providing a standardized and quantitative evaluation of individual PD motor characteristics. Such a system could improve clinical reliability and offer a deeper insight into specific movement abnormalities. 
	
	Recent advances in computer vision have led to new approaches for automating PD assessment~\cite{huckvale2019toward}. Video-based methods leveraging computer vision can offer a contactless, non-invasive and highly accessible solution~\cite{tarolli2020feasibility}. These methods could realize low-cost assessments using widely available devices, such as smartphones or laptop cameras, and allow for a standardized and frequent evaluation of the finger-tapping test~\cite{larson2021new}. However, existing approaches typically focus on reproducing imperfect clinical evaluations using black-box machine learning models. Thus, they lack clinical interpretability and do not provide insights into the individual motor characteristics ~\cite{li2021automated}. Moreover, most of these approaches have been developed using small video datasets only~\cite{williams2020supervised, deng2024interpretable}, which had been recorded under highly controlled conditions~\cite{yu2023clinically, khan2014computer}. As a result, it remains unclear how well the reported assessment performances will transfer to real-world clinical- or even home environments. 
	
	In recent years, several studies have explored methods to quantify PD severity using finger-tapping tests~\cite{skaramagkas2023multi,amo2024computer}. They can be categorized into two groups: (1) methods focusing on specific motor characteristics and (2) methods focusing on MDS-UPDRS score classification. Approaches in the first category have sought to isolate distinct PD characteristics. For example, Zhao et al.~\cite{zhao2020time} applied clustering to time-series signals to detect amplitude decrements in a finger-tapping tasks. Similarly, Heye et al.~\cite{heye2024validation} focused on extracting tapping frequency, amplitude, and rhythm variability to quantify bradykinesia and hesitation-halts. In contrast, the second group aimed to classify MDS-UPDRS scores without explicitly categorizing features into distinct groups of motor deficits. For instance, Islam et al.~\cite{islam2023using} used the Mediapipe~\cite{lugaresi2019mediapipe} model to identify three key points (wrist, thumb tip, and index finger positions) from each video frame. Two lines are drawn from wrist to thumb and wrist to index finger and the angle between these two lines is computed. From this angular displacement signal, they derived features and trained a light gradient boosting machine (LightGBM) to classify MDS-UPDRS scores across five severity levels. Although it achieves high accuracy, the high feature dimensionality (116 features) without clear grouping for distinct motor characteristics, as proposed by Islam and colleagues, challenges the clinical interpretability. Similarly, Guarin et al.~\cite{guarin2024characterizing} leveraged angular displacements as features, and applied logistic regression to classify videos into four MDS-UPDRS severity levels. Guo et al.~\cite{guo2022vision} used depth video to capture 3D hand movement representations. Hand regions were identified using the you only look once (YOLO) object detection model~\cite{redmon2018yolov3}.  Thumb and index finger key points were used to extract features such as amplitude, velocity, and frequency. A support vector machine (SVM) was trained on these features to classify MDS-UPDRS scores across five severity levels. Although depth cameras provide more precise representation, the higher cost and limited availability of depth cameras make this approach impractical for home-based monitoring. Deng et al.~\cite{deng2024interpretable} expanded previous methods by analyzing all fingers (including middle, little, and ring fingers) and calculated first- and second-order derivatives to obtain velocity and acceleration metrics for each finger. However, their classification was limited to only two MDS-UPDRS severity levels. 
	
	Several studies~\cite{lu2021quantifying,yang2024fasteval,li2021automated} have explored deep learning models to estimate MDS-UPDRS scores. Williams et al.~\cite{williams2020supervised} used convolutional neural networks (CNNs) combined with optical flow to extract motion features and classify MDS-UPDRS scores into two levels. Yang et al.~\cite{yang2022automatic} trained a small deep neural network (DNN) using tapping rate, frozen time, and amplitude variation to predict four MDS-UPDRS severity levels. Raw hand key points were fed into a multi-channel 1D CNN to capture temporal patterns at different scales and predict MDS-UPDRS scores~\cite{yang2024fasteval}. More advanced CNN-based models have been proposed, such as the three-stream architecture by Lu et al.~\cite{lu2021quantifying}, which processes joint distance, slow-motion, and fast-motion features for severity classification. A related study by Lu et al.~\cite{li2021automated} designed a three-stream network incorporating pose, motion, and geometric features, and leveraged a 1D CNN for MDS-UPDRS score classification. These approaches have demonstrated high accuracy. However, as they primarily aim to replicate MDS-UPDRS scores, they remain constrained by the subjectivity of the score. Instead, the broader aim is to leverage objective, video-based features to achieve a more granular and interpretable understanding of individual motor characteristics in PD. This could enable earlier detection of subtle impairments, better monitoring of symptom progression, and a more personalized assessment of treatment effects. While correlating with existing clinical scores remains essential to confirm their validity, the true potential of video-based assessment lies in surpassing the limitations of subjective ratings by offering consistent, quantitative, and clinically meaningful insights.
	
\begin{figure*}[ht!]
		\centering
		\includegraphics[width=0.99\textwidth]{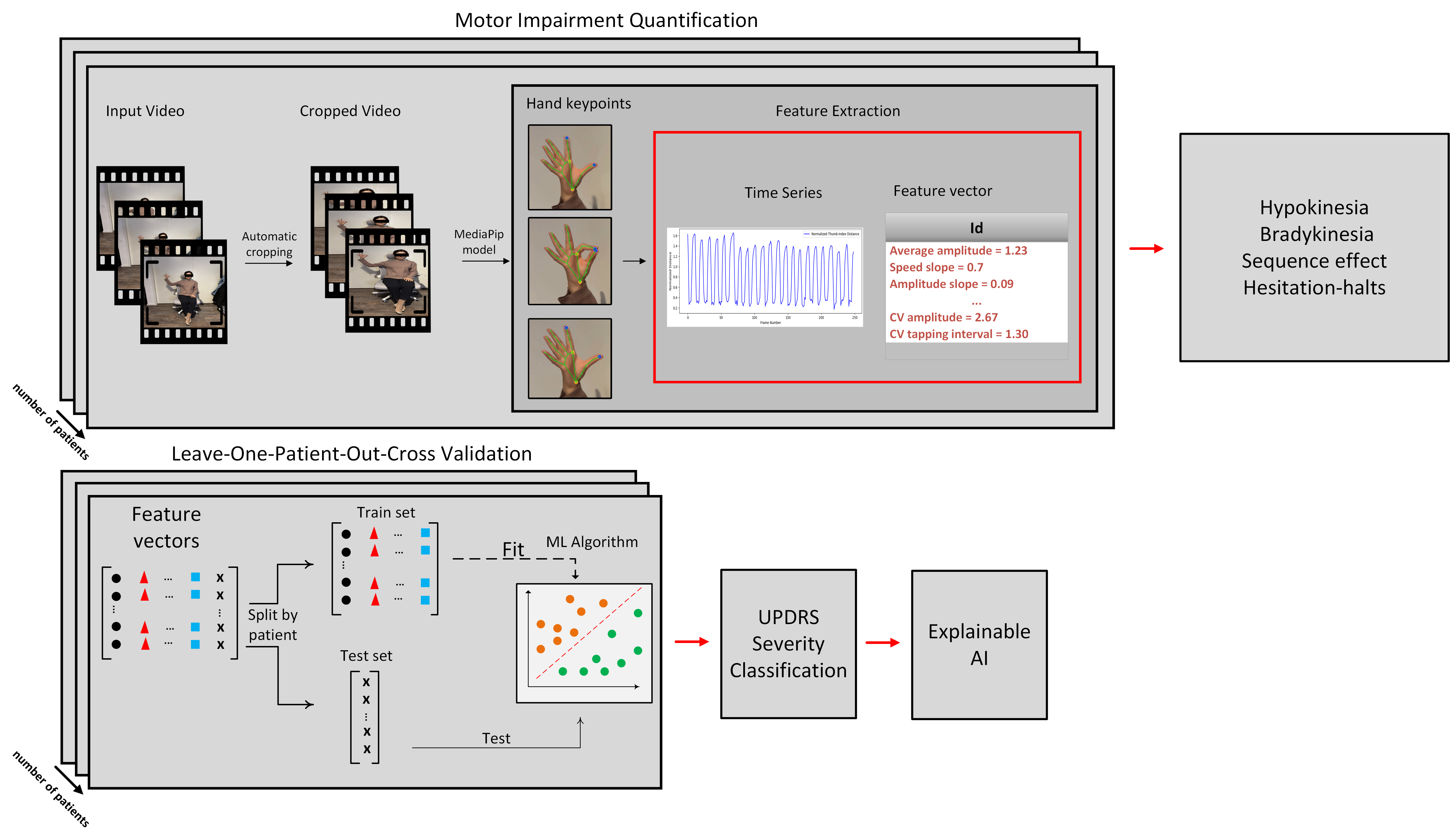}
		\caption{\textbf{Block diagram describing  the analysis pipline of the proposed method.} The system detects key points—including the index finger, thumb, and wrist—in each video frame and tracks them throughout the video sequence. Next, a time-series signal is generated, representing the distance between the thumb and index finger, which reflects the patient’s ability to fully open their fingers. This signal is scaled by the palm length to account for variations in the patient’s distance from the camera. Features are extracted from the signal to assess PD severity across four main domains: hypokinesia, bradykinesia, sequence effect, and hesitation/halts. Per video, the MDS-UPDRS score values are then estimated  by machine learning classifiers, which had been trained on these features. The classification performances are estimated using a leave-one-patient-out cross-validation scheme. Finally, the trained classifiers are introspected.}
		\label{fig:single_figure}
	\end{figure*}

	To address these challenges, this paper presents a novel video-based framework for quantifying individual, clinically relevant PD motor characteristics from finger-tapping recordings. The block diagram in Figure~\ref{fig:single_figure} provides an overview of the proposed processing pipeline. First, the input video is pre-processed to extract hand key points using Mediapipe~\cite{lugaresi2019mediapipe}. Using these, a time-series signal representing the distance between the thumb tip and the index finger tip is generated. The resulting distance values are used to extract clinically interpretable features, which allow for quantification of individual motor characteristics, based on a recently updated clinical definition of bradykinesia~\cite{bologna2023redefining}. We apply principal component analysis (PCA) with varimax rotation to assess whether these video-based features co-vary according to the four proposed clinical motor characteristics, and to reveal potential additional components. To verify the clinical validity of our approach, we evaluate whether these features can be used to train machine learning classifiers to estimate the MDS-UPDRS finger-tapping score. Our approach is validated using a relatively large-scale dataset of 485 videos derived from 74 patients.
	
	\begin{table}[ht!]
		\small
		\caption{MDS-UPDRS scoring criteria for finger tapping test}
		\label{table:mds_MDS-UPDRS}
		\centering
		\resizebox{\columnwidth}{!}{ 
			\begin{tabular}{p{1cm}p{7cm}}
				\hline
				Score & Scoring Criteria \\
				\hline
				0 & Without any problem \\
				\hline
				1 & Any of the following: \\
				& a) The regular rhythm is broken with one or two \textbf{interruptions} or hesitations of the tapping movement; \\
				& b) Slight \textbf{slowing}; \\
				& c) The \textbf{amplitude decrements} near the end of the 10 taps \\
				\hline
				2 & Any of the following: \\
				& a) 3 to 5 \textbf{interruptions} during tapping; \\
				& b) Mild \textbf{slowing}; \\
				& c) The \textbf{amplitude decrements} midway in the 10-tap sequence \\
				\hline
				3 & Any of the following: \\
				& a) More than 5 \textbf{interruptions} or at least one longer freeze in ongoing movement; \\
				& b) Moderate \textbf{slowing}; \\
				& c) The \textbf{amplitude decrements} starting after the 1st tap \\
				\hline
				4 & Cannot or can only barely perform the task because of \textbf{slowing, interruptions, or decrements} \\
				\hline
			\end{tabular}
		}
	\end{table}
	
	\section*{Results}
    To demonstrate the clinical validity and practical value of our approach, we analyzed the extracted features from multiple perspectives. First, we assessed whether the features align with expert-rated MDS-UPDRS scores to confirm that they capture clinically relevant patterns. Next, we used PCA with varimax rotation to explore the underlying structure of motor characteristics and to validate or refine clinical groupings. We then evaluated how well the features support automated classification of MDS-UPDRS scores using multiple machine learning models. Finally, we benchmarked our method against state-of-the-art approaches.
	\subsection*{Comparison of video-based features and clinical evaluation}
\begin{figure*}[ht!]
    \centering
	\includegraphics[width=1\textwidth]{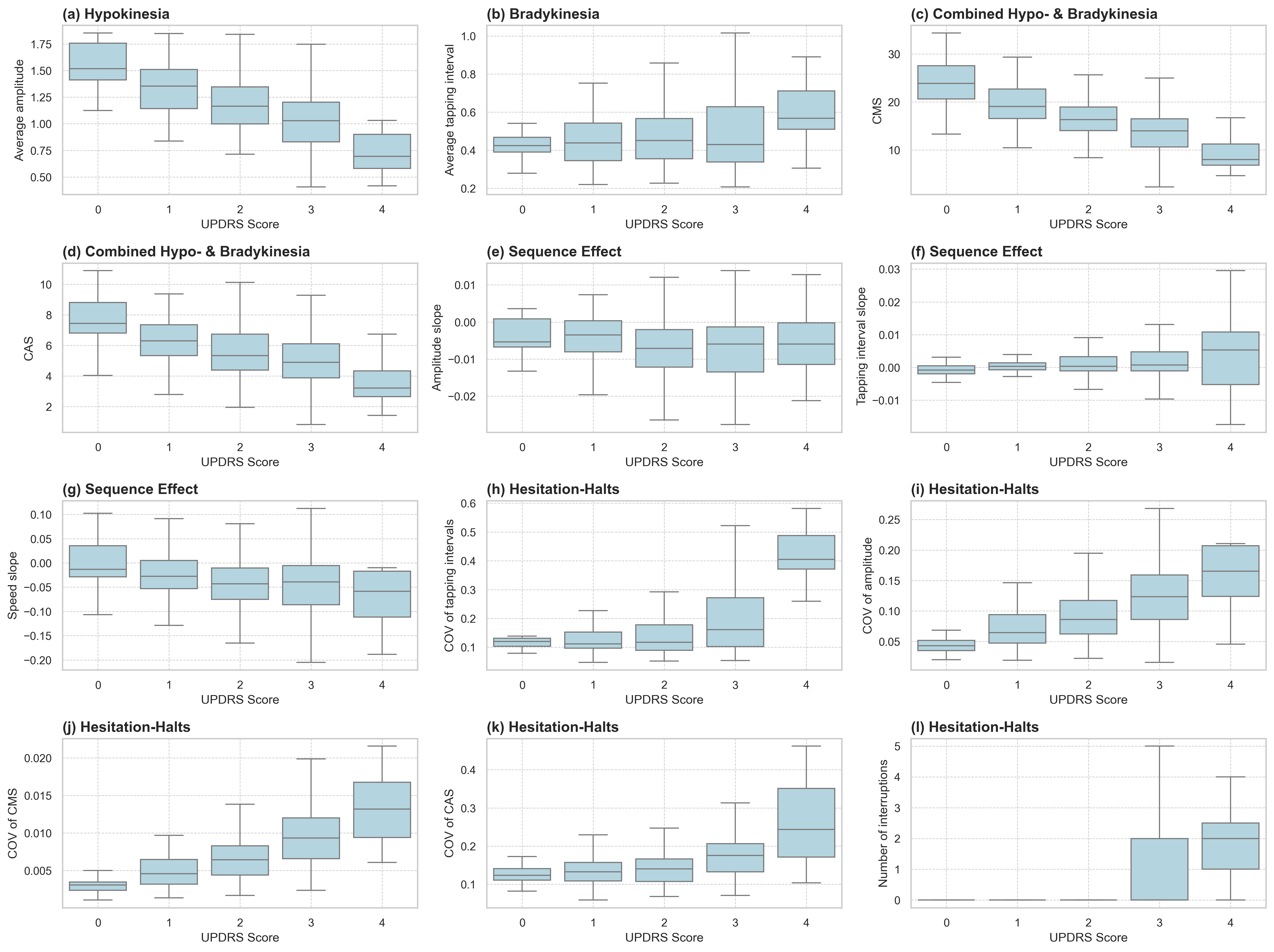}
	\caption{\textbf{Distribution of clinically interpretable features across MDS-UPDRS finger-tapping scores.} 
Each subplot shows the distribution of a specific feature across MDS-UPDRS scores (0–4), highlighting its relation to motor impairment severity. 
(a) Hypokinesia: average amplitude, measuring hand opening ability. 
(b) Bradykinesia: Average tapping interval. 
(c, d) Combined hypo- \& bradykinesia: CAS and CMS. (e-g) Sequence effect: amplitude decrement, tapping interval increment, speed decrement. (h–l) Hesitation-halts: COV in tapping interval, amplitude, CMS, CAS, and number of interruptions. Each feature reflects a clinically relevant aspect of motor impairment, with distributions aligning with increasing disease severity}
	\label{fig:box_plots}
\end{figure*}

	To establish the clinical validity of the extracted video-based features, we first assessed how well they align with clicial MDS-UPDRS finger-tapping scores. This analysis determines whether the proposed features meaningfully capture the motor impairments typically observed during clinical assessments. We quantified 10 features measuring four groups of finger tapping motor characteristics: hypokinesia, bradykinesia, hesitations-halts, and the sequence effect. Also, 2 other features capture the combined effect of hypokinesia and bradykinesia. The distribution of these features across MDS-UPDRS scores of the finger tapping task (0 to 4) is presented in Figure~\ref{fig:box_plots}. 
    
\begin{figure*}[ht!]
		\centering
		\begin{subfigure}{0.45\textwidth}
			\centering
			\includegraphics[width=\textwidth]{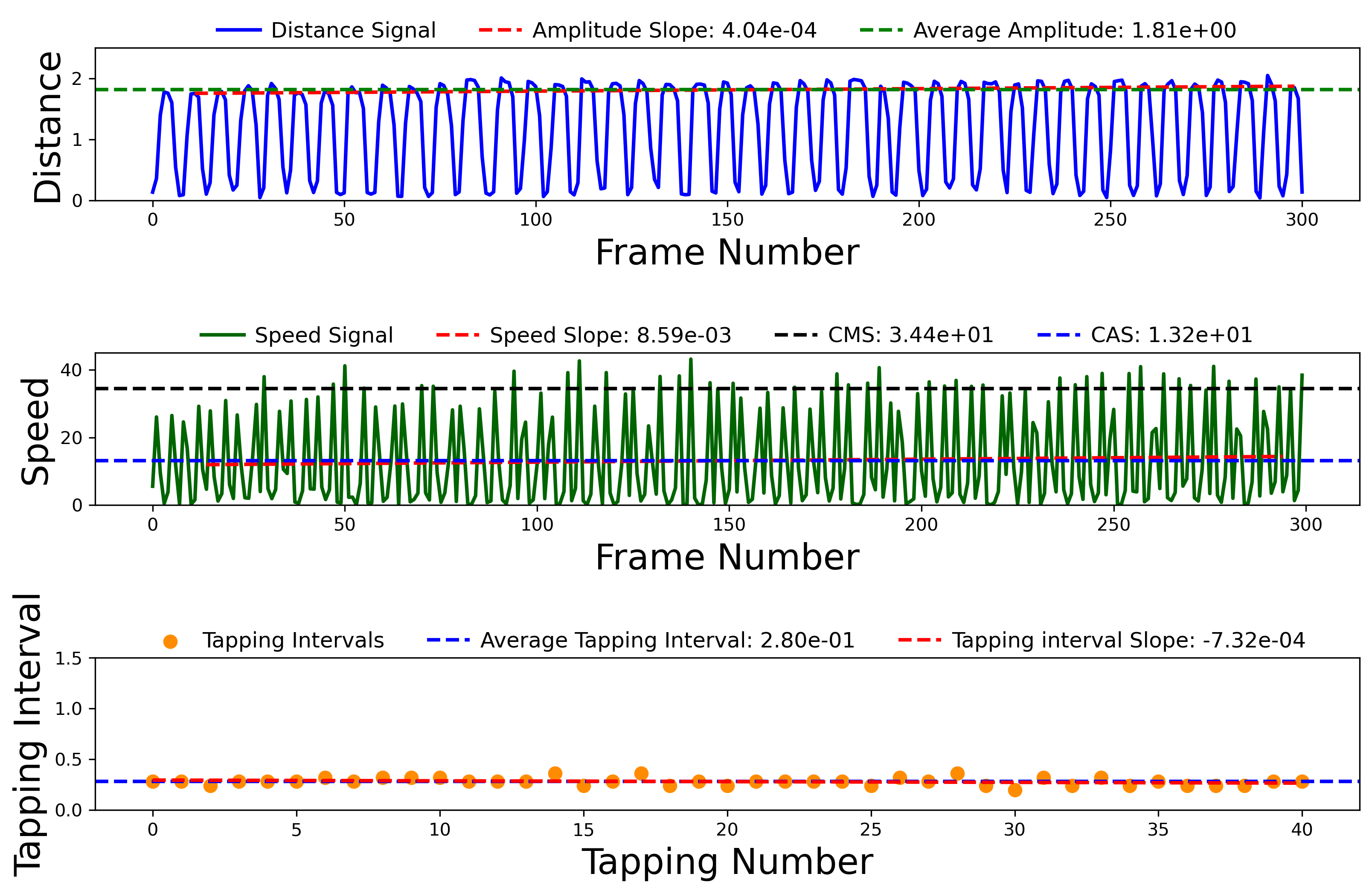}
			\caption{Patient with MDS-UPDRS score of 0}
			\label{fig:feature_signal_0}
		\end{subfigure}
		\hfill
		\begin{subfigure}{0.45\textwidth}
			\centering
			\includegraphics[width=\textwidth]{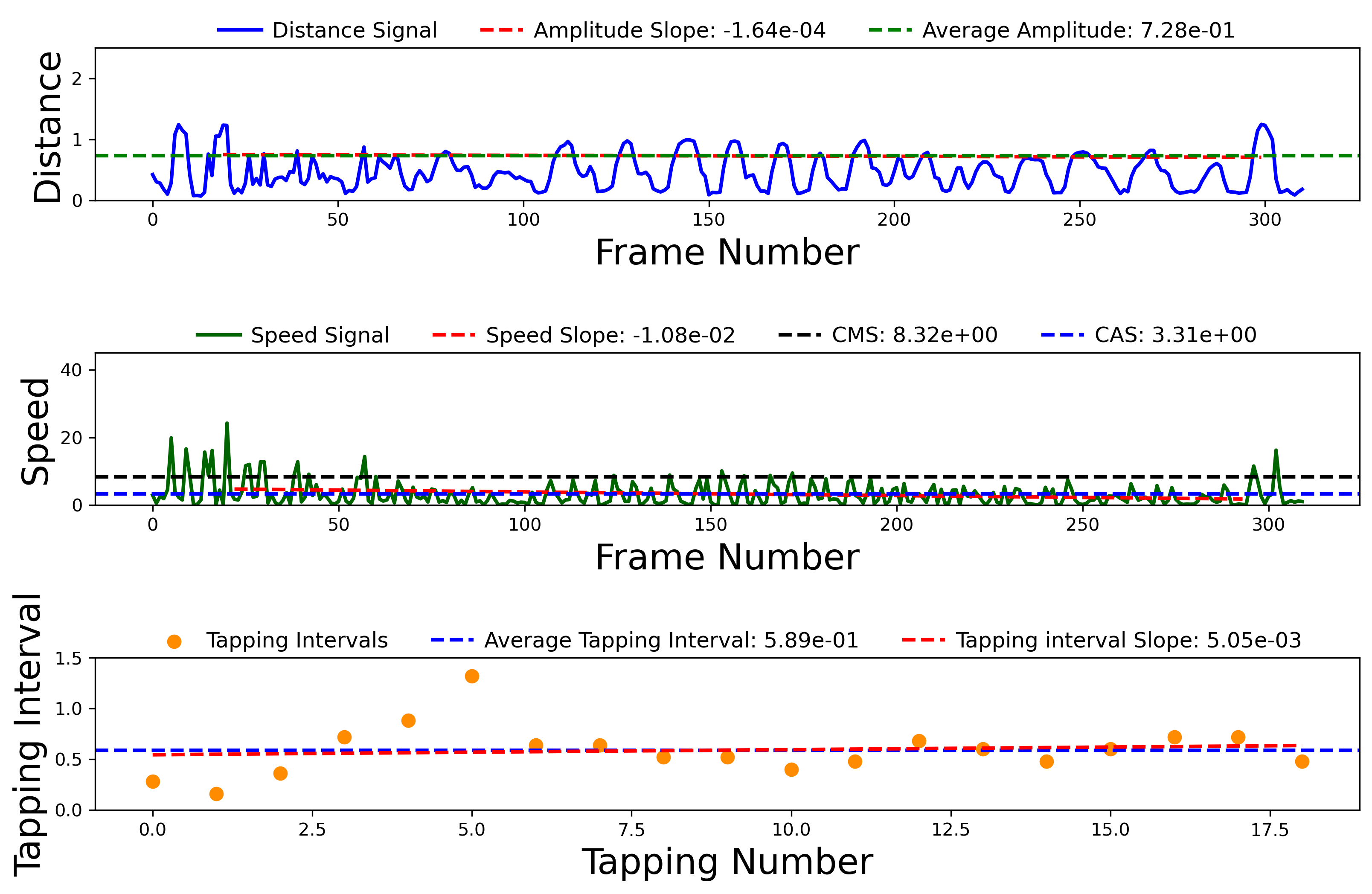}
			\caption{Patient with MDS-UPDRS score of 4}
			\label{fig:feature_signal_3}
		\end{subfigure}
		\caption{\textbf{Amplitude, speed, and tapping interval for two patients with MDS-UPDRS scores of 0 and 4}. The distance signal for the patient with a MDS-UPDRS score of 0 demonstrates high-amplitude tapping, whereas the patient with a MDS-UPDRS score of 4 shows lower amplitudes. Similarly, the speed signal for the patient with a MDS-UPDRS score of 0 exhibits stable, high-speed tapping that highlights the ability of the patient to maintain a steady tapping throughout the task. In contrast, the speed signal for the patient with a MDS-UPDRS score of 4 reveals a significantly reduced tapping speed, which reduces over time. The tapping interval signals also reveal distinct patterns. The patient with a MDS-UPDRS score of 0 exhibits an average duration of 0.28 seconds. In contrast, the patient with a MDS-UPDRS score of 4 shows a prolonged average tapping interval of 0.56 seconds.}
		\label{fig:feat_visualization}
	\end{figure*}

    Subplot (a) focuses on the hypokinesia feature "average amplitude", which measures the extent to which patients can open their fingers during the task. This feature decreases as MDS-UPDRS scores increases, which reflects a reduced amplitude in patients with a higher disease severity. Subplot (b) presents the bradykinesia feature. The average tapping interval captures the time required to complete each tap, which is significantly higher for patients with a MDS-UPDRS score of 4. Cycle max speed (CMS) and cycle average speed (CAS) in subplots (c) and (d) capture both hypokinesia and bradykinesia, which progressively decrease as MDS-UPDRS scores increase. Subplots (e), (f) and (g) present the sequence effect features, which capture the change rates in amplitude, tapping interval and speed over time. As expected, higher MDS-UPDRS scores were associated with a higher decline of speed and higher increase of tapping interval. However, we did not find a stronger decline of amplitude in patients with higher MDS-UPDRS scores.Subplots (h) to (k) show the hesitation-halts features. The coefficients of variation (COVs) capture irregularity in amplitude, tapping interval and speed over time. Patients with more severe characteristics exhibit a larger variability in amplitude, tapping interval and speed. Lastly, the number of interruptions in subplot (l) also increases with disease severity. Together, these results show that all video-based features, except for the amplitude sequence effect feature, show the expected relationship with the clinical evaluation of the finger tapping task.
    Figures~\ref{fig:feature_signal_0} and~\ref{fig:feature_signal_3} show the distance and speed signals and features derived thereof for two illustrative examples: one patient with a MDS-UPDRS finger tapping score of 0 (minimal motor impairment), and one patient with a MDS-UPDRS score of 4 (severe motor impairment). The plots clearly show how amplitude, speed, and tapping interval differ between these two cases.
	
	\subsection*{Redefining motor characteristics categories based on principal component analysis with varimax rotation}
\begin{figure*}[ht!]
		\centering
		\includegraphics[width=0.8\textwidth]{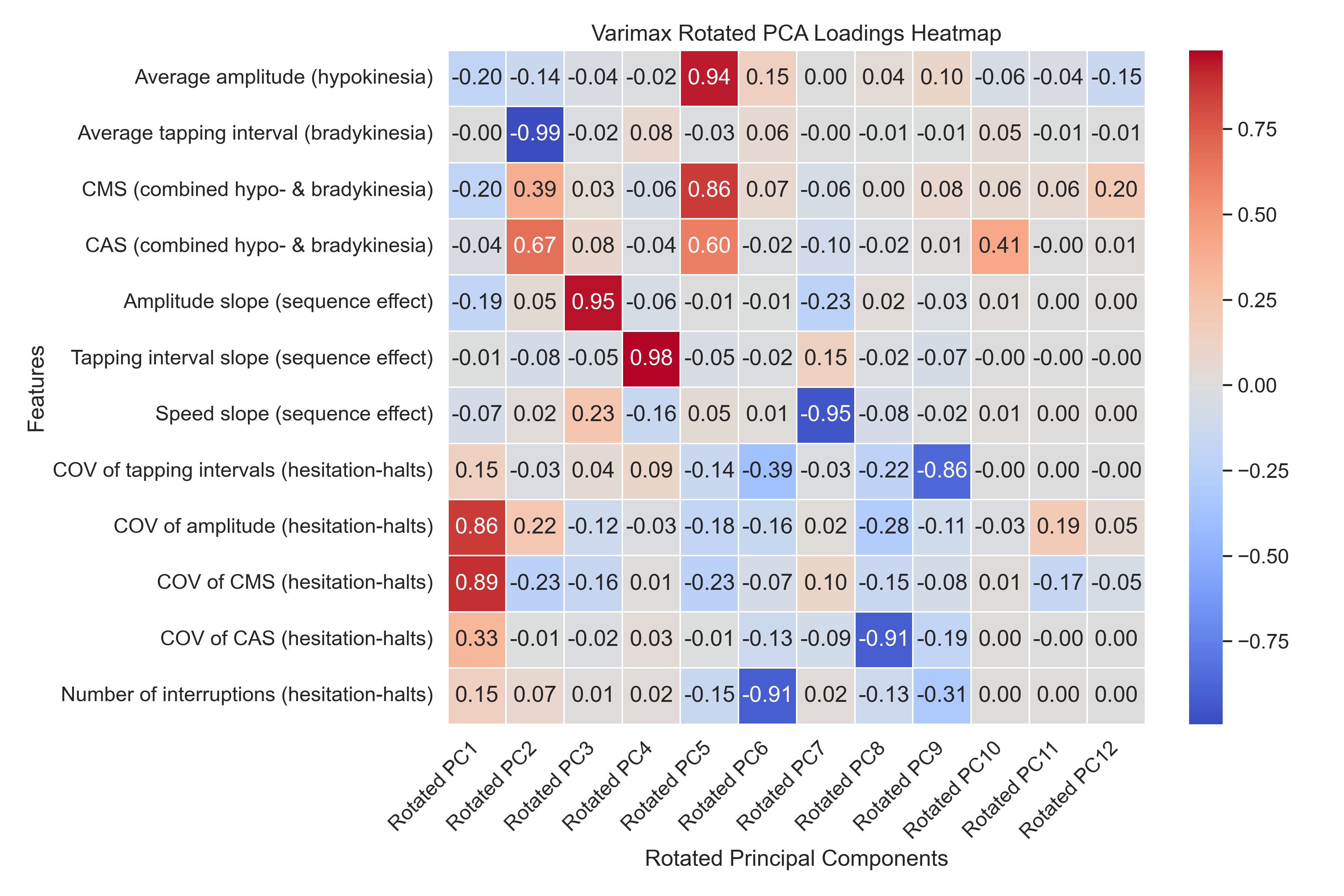}
		\caption{\textbf{PCA with varimax rotation for feature groupings and potential reclassification of motor deficits in PD}. Varimax rotation is applied to enhance interpretability by maximizing the separation between feature across components. The heatmap displays the loading values of each feature across the rotated principal components. Warmer colors (red) indicate strong positive loadings, and cooler colors (blue) represent strong negative loadings. Features with high loadings on the same component suggest shared variance and likely represent related motor characteristics.}
		\label{fig:rotated}
	\end{figure*}

\begin{figure}[ht!]
		\centering
		\includegraphics[width=0.5\textwidth]{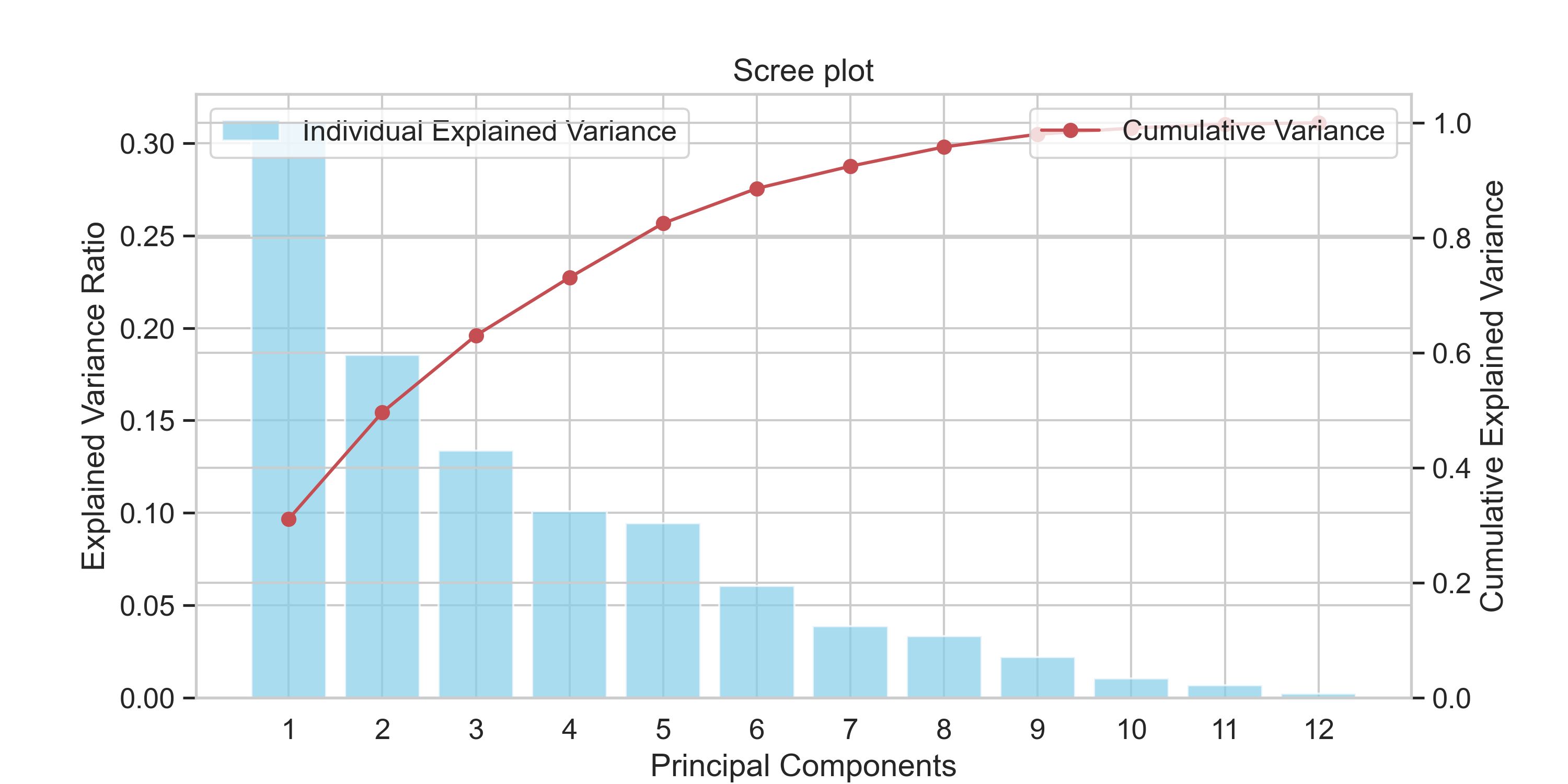}
        \caption{\textbf{Scree plot of principal components.} The blue bars represent the individual explained variance ratio for each principal component, indicating how much variance in the data is captured by that component. The red line shows the cumulative explained variance, which represents the total amount of variance accounted for when combining the first n components. This plot demonstrates that the first 9 components together explain approximately 98\% of the total variance.}
		\label{fig:variance}
	\end{figure}

	To evaluate the validity of the categorization of motor deficits proposed by Bologna et al.~\cite{bologna2023redefining}, we applied PCA with varimax rotation~\cite{kaiser1958varimax} to the feature set. Varimax rotation was applied to enhance interpretability by producing sparse loadings~\cite{kaiser1958varimax}. The goal was to assess whether the hypothesized four motor deficit groups — hypokinesia, bradykinesia, sequence effect, and hesitation-halts — emerge from the data. Figure~\ref{fig:rotated} depicts the component loading heatmaps. Nine principal components (PCs) were selected for further analysis, because up to the ninth component, each PC was primarily characterized by distinct features (i.e., from the tenth component onward, features with high loadings were already prominently represented in earlier components). The first nine components retained 98\% of the explained variance (see Figure~\ref{fig:variance} for the explained variance of all components). The results in Figure~\ref{fig:rotated} revealed that PC1 mainly captures features associated with irregularity in amplitude and speed. PC2 is characterized by average tapping interval (bradykinesia) and combined hypo-\& bradykinesia. PC3 and PC4 capture different aspects of the sequence effect, suggesting that amplitude decrement and tapping interval increment may represent separate underlying patterns. PC5 corresponds to hypokinesia and combined hypo-/bradykinesia features. PC6 is dominated by the number of interruptions, which suggests that abrupt pauses may be treated as a separate motor deficit. PC7 includes speed sequence effect and finally, PC8 and PC9 captured irregularity in tapping time intervals and speed. On the one hand, the identified components partly overlap with the four clinical motor deficits. However, the results suggest that both sequence effect and hesitation-halts may not be single constructs. Specifically, sequence effect appears to involve three distinct components—amplitude decrement, speed decrement, and progressive increase in tapping interval—while hesitation-halts decomposes into separate components for irregularities in amplitude, speed, and timing. These findings support a more fine-grained characterization of PD motor characteristics.
    
	\subsection*{MDS-UPDRS score classification}
	To verify the clinical validity of our approach, we evaluate whether the video-based features can be used to train machine learning classifiers to estimate the MDS-UPDRS finger-tapping score. The classification performance metrics, including accuracy, balanced accuracy, acceptable accuracy, macro precision, and macro F1-score, along with their confidence intervals, are presented in Table~\ref{tab:classification_metrics_ci}. Differences between the different classifiers (random forest, logistic regression and LightGBM) and between multi-class and ordinal classification were generally small (all 95\% confidence intervals were overlapping). We also compared our models against two baselines: a random guess baseline, which assigns each class an equal probability, and a majority class baseline, which always predicts the most frequent class. Our proposed models significantly outperform both baselines across all evaluation metrics. Because the ordinal random forest classifier demonstrated a slightly higher performance on multiple metrics, this model was selected for further analysis in the following sections.  
	
	\begin{table*}[ht!]
		\small
		\centering
		\caption{\textbf{Performance of different classifiers.} The first three rows show the results of logistic regression, LightGBM, and random forest models trained on our extracted features. The reported values represent the mean performance across all folds in leave-one-patient-out cross-validation, with 95\% confidence intervals shown in brackets. The \textit{Random Guess Baseline} refers to a classifier that assigns videos to 5 classes at random, while the \textit{Majority Class Baseline} corresponds to a classifier that always predicts the most frequent class in the dataset}
		\label{tab:classification_metrics_ci}
		\resizebox{\textwidth}{!}{%
			\begin{tabular}{@{}lcccccccccc@{}}
				\toprule
				\multirow{2}{*}{Metrics (\%)} & \multicolumn{2}{c}{Acc} & \multicolumn{2}{c}{Balanced Acc} & \multicolumn{2}{c}{Acceptable Acc} & \multicolumn{2}{c}{Macro Precision} & \multicolumn{2}{c}{Macro F1} \\
				\cmidrule(lr){2-3} \cmidrule(lr){4-5} \cmidrule(lr){6-7} \cmidrule(lr){8-9} \cmidrule(lr){10-11}
				& Multi & Ordinal & Multi & Ordinal & Multi & Ordinal & Multi & Ordinal & Multi & Ordinal \\
				\midrule
				Logistic regression & \(54.05\,[47.34–60.75]\) & \(54.27\,[47.34–60.75]\) & \(48.65\,[42.33–54.97]\) & \(48.38\,[42.10–54.66]\) & \(\mathbf{95.84\,[92.62–99.07]}\) & \(95.69\,[92.45–98.94]\) & \(41.34\,[34.49–48.18]\) & \(41.55\,[34.87–48.23]\) & \(40.67\,[34.25–47.09]\) & \(40.25\,[33.99–46.51]\) \\
				LightGBM & \(51.45\,[44.35–58.56]\) & \(54.20\,[47.84–60.57]\) & \(46.03\,[39.24–52.82]\) & \(49.20\,[42.70–55.71]\) & \(92.53\,[88.06–97.01]\) & \(93.32\,[89.61–97.03]\) & \(38.65\,[31.67–45.63]\) & \(42.48\,[36.04–48.92]\) & \(38.98\,[32.21–45.75]\) & \(40.96\,[34.78–47.14]\) \\
				Random forest & \(54.03\,[46.98–61.08]\) & \(\mathbf{55.95\,[49.31–62.60]}\) & \(48.30\,[41.44–55.17]\) & \(\mathbf{49.65\,[43.21–56.08]}\) & \(93.93\,[90.46–97.41]\) & \(94.06\,[90.64–97.48]\) & \(\mathbf{45.12\,[38.08–52.16]}\) & \(44.37\,[37.39–51.35]\) & \(\mathbf{42.99\,[36.11–49.87]}\) & \(42.21\,[35.53–48.89]\) \\
				
				Random guess baseline & \multicolumn{2}{c}{20.00} & \multicolumn{2}{c}{20.00} & \multicolumn{2}{c}{58.79} & \multicolumn{2}{c}{20.00} & \multicolumn{2}{c}{20.00} \\
				Majority class baseline & \multicolumn{2}{c}{45.15} & \multicolumn{2}{c}{20.00} & \multicolumn{2}{c}{84.12} & \multicolumn{2}{c}{9.03} & \multicolumn{2}{c}{12.44} \\
				
				\bottomrule
			\end{tabular}
		}
	\end{table*}
	
	\subsection*{SHAP method for interpreting the classifier output}
\begin{figure}[ht!]
		\centering
		\includegraphics[width=0.5\textwidth]{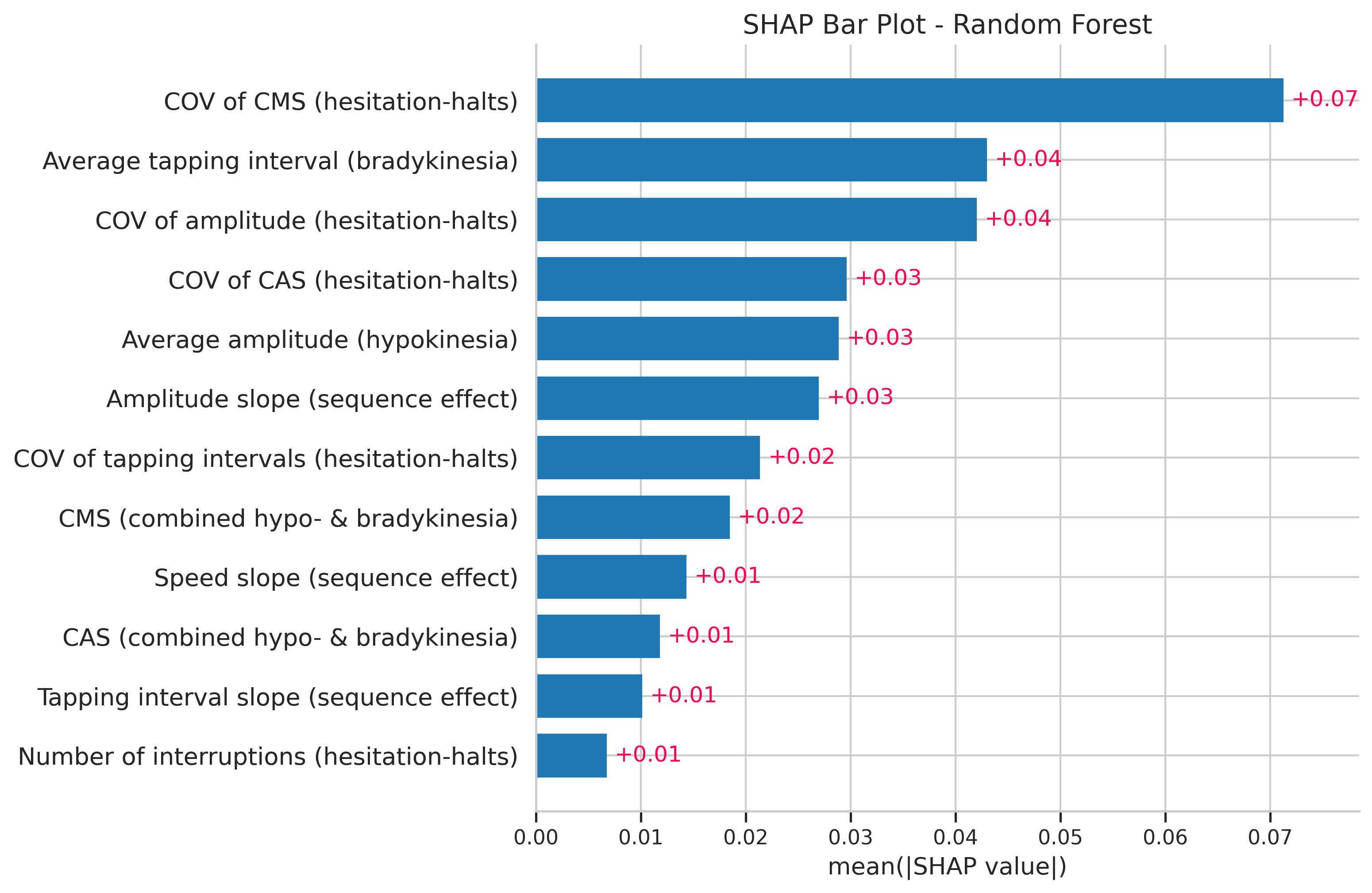} 
		\caption{\textbf{Feature importance based on SHAP values for the random forest classifier.} The bar plot summarizes the global feature importance using SHAP values. The x-axis represents the mean absolute SHAP value for each feature, which quantifies how much, on average, that feature contributes to the model's predictions. Features are sorted in descending order of importance. A higher SHAP value indicates that the feature has a greater overall impact on the model’s output.}
		\label{fig:shap}
	\end{figure}

	To understand the contribution of each feature to the classifier’s predictions, we employed SHAP, an explainable AI method that quantifies the marginal impact of each feature on model predictions. SHAP achieves this by excluding features from subsets and aggregating their influence on model output. Figure~\ref{fig:shap} presents the SHAP plot for the random forest classifier and illustrates the importance of each feature in predicting MDS-UPDRS scores. Higher SHAP values indicate a stronger influence on the model’s decision-making.  The results show that hesitation-halts features including COV of CMS, COV of amplitude, and COV of CAS have the highest absolute SHAP values. Average tapping intervals (bradykinesia) also contributes significantly to the model's predictions. In contrast, sequence effect features, including speed decrement and tapping interval increment, exhibit moderate impact on classification. Finally, the number of interruptions shows the lowest SHAP value which indicates that they contribute less to MDS-UPDRS classification compared to other features. 
	
	\subsection*{Comparison with related works for MDS-UPDRS finger-tapping score classification}
\begin{figure*}[ht!]
		\centering
		\includegraphics[width=1\textwidth]{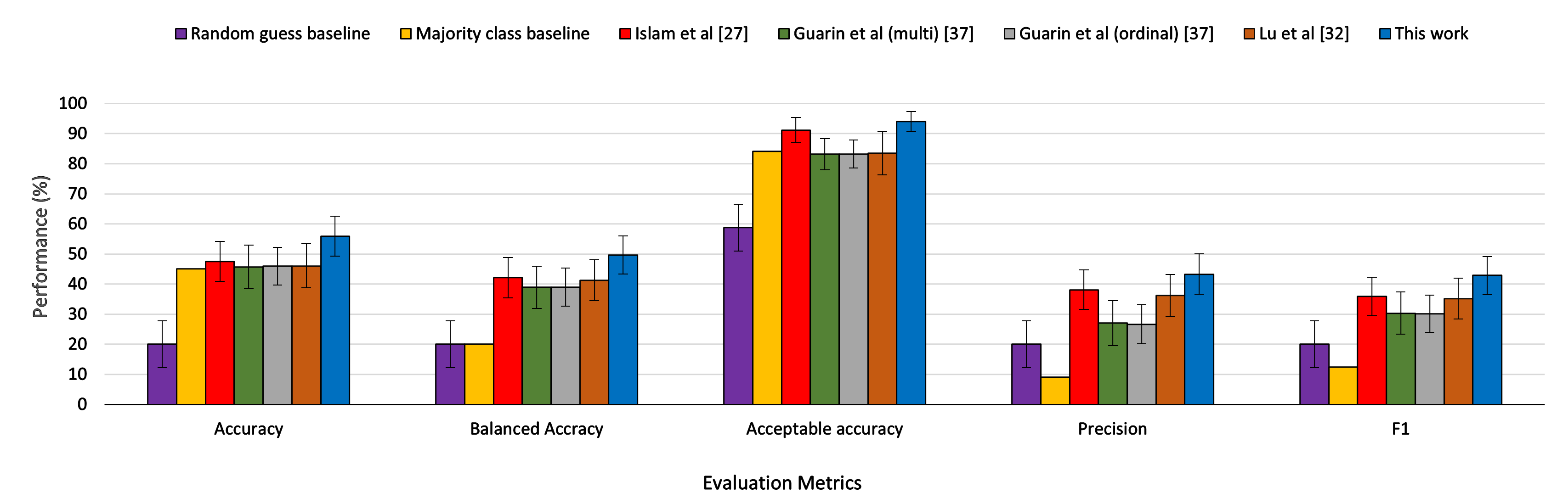} 
		\caption{\textbf{Comparison of the proposed method with existing state-of-the-art approaches for MDS-UPDRS finger-tapping score classification.} Results for the method by Islam et al.~\cite{islam2023using} are obtained by using the official implementation from the authors' GitHub repository. To ensure fair comparison, their hyperparameters are re-tuned on our dataset using the optuna library. For Guarin et al.~\cite{guarin2024trained} and Lu et al.~\cite{lu2021quantifying}, official codes are not publicly available; therefore, these methods are re-implemented based on their publications. All methods are trained and evaluated on our dataset under identical experimental settings using leave-one-patient-out cross-validation. Performance metrics include accuracy, balanced accuracy, acceptable accuracy, macro precision, and macro F1-score. Bars represent the mean values across all test folds, with 95\% confidence intervals shown as error bars.}
		\label{fig:comparison}
	\end{figure*}

    Figure~\ref{fig:comparison} presents a detailed comparison of our MDS-UPDRS 
 classification method with state-of-the-art approaches. Our proposed method consistently outperforms related works across all evaluated metrics. Our method achieves an accuracy of 55.95\%, a significant improvement of 17.76\% over the highest accuracy of 47.51\% achieved by  Islam et al (p=0.002). Our method achieves a balanced accuracy of 49.65\%, which represents an improvement of 17.81\% over 42.14\% achieved by Islam et al.\cite{islam2023using} (p=0.004). Also, the acceptable accuracy of 94.06\%, macro precision of 43.29\% and macro F1-score of 42.87\%, are higher than all other approaches.
	
	\subsection*{Comparison of distance-based and angle-based signals}
\begin{figure*}[ht!]
		\centering
		\includegraphics[width=1\textwidth]{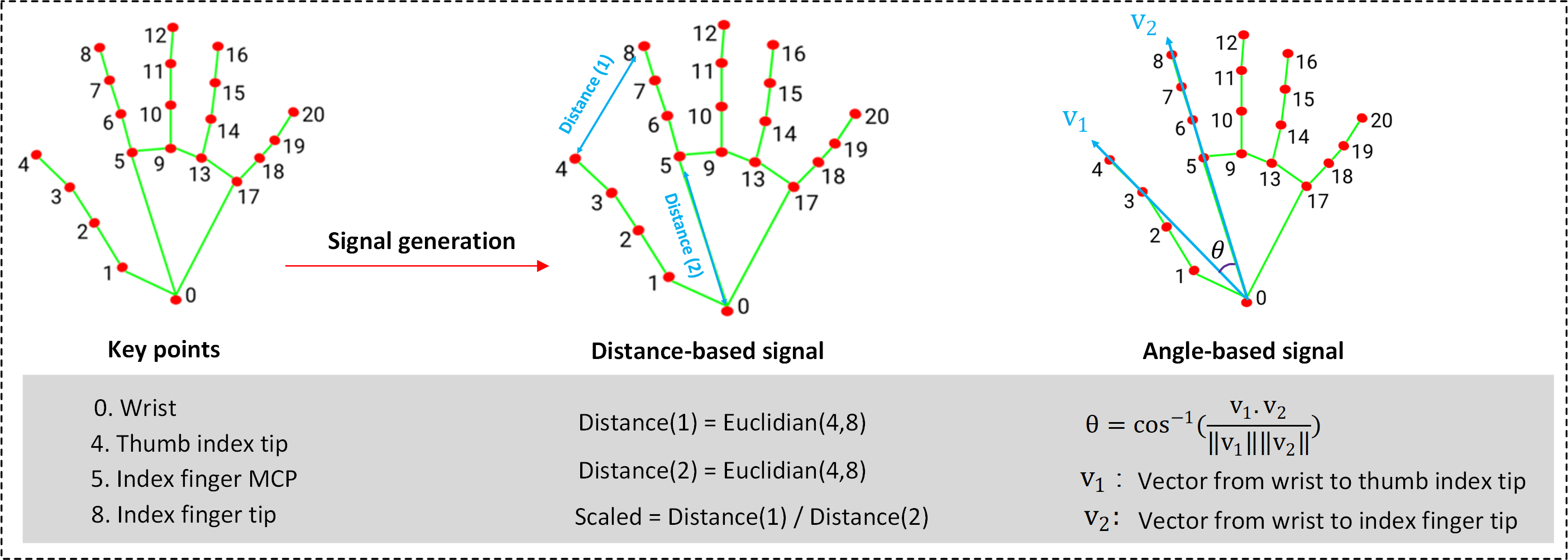} 
		\caption{\textbf{Signal generation based on the distance between the thumb and index finger (middle) and the angle between the thumb and index finger (right)}. The distance-based signal is calculated as the Euclidean distance between the thumb tip and index finger tip divided by palm length. The angle-based signal computed as the angle formed between vectors from the wrist to the thumb and wrist to the index finger.}
		\label{fig:signal_generation}
	\end{figure*}

\begin{figure}[ht!]
		\centering
		\includegraphics[width=0.5\textwidth]{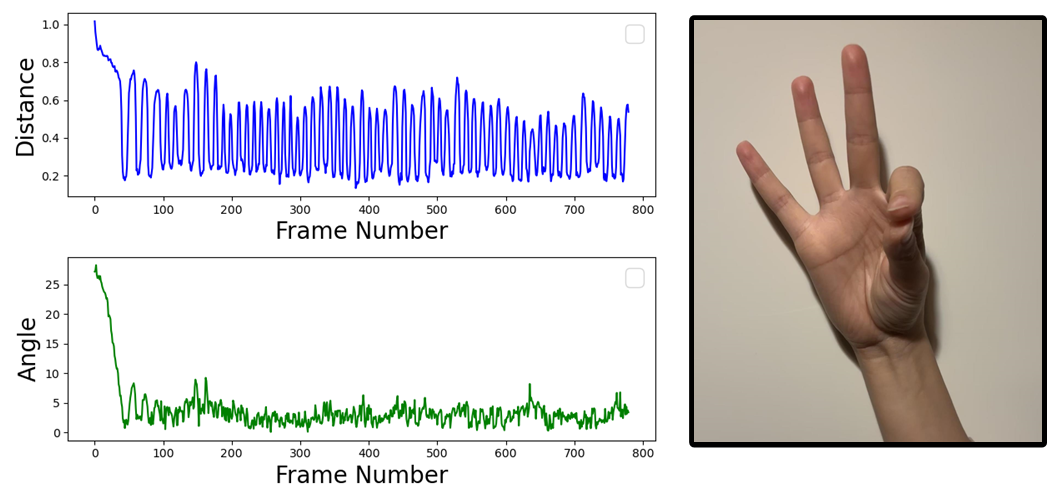} 
		\caption{\textbf{An example of a camera viewpoint effect where angle-based signals (green) are less accurate compared to distance-based signals (blue)}. The wrist, thumb, and index finger are nearly aligned in a straight line during the tapping task. Consequently, the vectors used in angle computation are overlapping, resulting in an angle close to zero throughout the sequence. This leads to a flat or minimally varying angular signal that fails to reflect the dynamics of finger tapping. In contrast, the distance-based signal remains sensitive to variations in hand opening and closing.}
		\label{fig:angle_dis}
	\end{figure}

	To understand what explains the higher performance of our approach compared to existing models, we assessed the impact of differences between the different pipelines. An important difference with existing models is that our signal representation leverages distance-based signals, which are less sensitive to variations in camera viewpoints compared to angle-based representations~\cite{islam2023using}~\cite{guarin2024trained}. The difference between distance- and angle-based signals is illustrated in figure~\ref{fig:signal_generation}. Figure~\ref{fig:angle_dis} shows an example filmed from the front view, where the distance-based signal (blue plot) provide a more robust representation of the finger-tapping movement compared to the angle-based signals (green plot).
	To further evaluate the impact of using distance-based signals, we compared classifier performance between using distance- and angle-based features. All classifier parameters were optimized using the same procedures as described in~\ref{sec:classification}. The results, summarized in Table~\ref{tab:classification_metrics_angle_dis}, demonstrate that distance-based features consistently outperform angle-based features across all classifiers and evaluation metrics.\\

\begin{table*}[ht!]
	\small
	\centering
	\caption{\textbf{Comparison of classification metrics using distance-based and angle-based features.} Three classifiers (logistic Regression, LightGBM, and random forest) are trained on two different input modalities: distance signal and angle signal. The same feature extraction pipeline was applied separately to the distance and angle signals. Reported values represent the mean performance across all folds in leave-one-patient-out cross-validation, with 95\% confidence intervals shown in brackets.}
	\label{tab:classification_metrics_angle_dis}
	\resizebox{\textwidth}{!}{%
		\begin{tabular}{@{}llcccccc@{}}
			\toprule
			Features & Classifier & Accuracy (\%) & Balanced Accuracy (\%) & Acceptable Accuracy (\%) & Macro Precision (\%) & Macro F1 (\%) \\
			\midrule
			Distance-Based & Logistic regression & \(\mathbf{54.27 [47.63, 60.90]}\) & \(\mathbf{48.38 [42.10, 54.66]}\) & \(\mathbf{95.69 [92.45, 98.94]}\) & \(\mathbf{41.55 [34.87, 48.23]}\) & \(\mathbf{40.25 [33.99, 46.51]}\) \\
			Angle-Based & Logistic regression & \(46.69 [39.76, 53.63]\) & \(42.11 [35.63, 48.60]\) & \(85.88 [80.30, 91.46]\) & \(34.83 [28.31, 41.35]\) & \(34.00 [27.81, 40.20]\) \\
			\midrule
			Distance-Based & LightGBM & \(\mathbf{54.20 [47.84, 60.57]}\) & \(\mathbf{49.20 [42.70, 55.71]}\) & \(\mathbf{93.32 [89.61, 97.03]}\) & \(\mathbf{42.82 [36.04, 48.92]}\) & \(\mathbf{40.96 [34.78, 47.14]}\) \\
			Angle-Based & LightGBM & \(48.47 [41.59, 55.35]\) & \(42.79 [36.46, 49.12]\) & \(90.09 [84.72, 96.47]\) & \(37.20 [30.17, 44.24]\) & \(35.77 [29.28, 21.26]\) \\
			\midrule
			Distance-Based & Random forest & \(\mathbf{55.95 [49.31, 62.60]}\) & \(\mathbf{49.65 [43.21, 56.08]}\) & \(\mathbf{94.06 [90.64, 97.41]}\) & \(\mathbf{43.29 [36.36, 50.23]}\) & \(\mathbf{42.87 [36.14, 49.59]}\) \\
			Angle-Based & Random forest & \(51.29 [44.63, 57.95]\) & \(47.51 [40.93, 54.08]\) & \(91.31 [87.28, 95.36]\) & \(41.74 [34.83, 48.65]\) & \(39.42 [32.71, 46.14]\) \\
			\bottomrule
		\end{tabular}
	}
\end{table*}

	\subsection*{Analysis of Misclassified samples}
	To better understand classification errors by our proposed model, we further analyze recordings where the model's predictions did not align with the clinical evaluation. 27 videos were misclassified by more than one distance from the clinical score. These videos were re-scored by a new assessor who was blind to both the previous assessor's score and the model-generated score. Out of these, 9 new clinical scores aligned with the model's prediction, 5 were aligned with the previous assessor's score, and 13 fell between the two.
    These findings indicate that among the misclassified samples, 33.33\% of the model-predicted scores were confirmed by the new assessor, suggesting that a portion of the model's misclassifications may be due to inherent variability in human scoring rather than a failure of the algorithm itself.

	\section*{Discussion}
	In this work, we adopt the updated clinical definition of bradykinesia to develop a granular, AI-based framework for quantifying motor impairments in PD. To the best of our knowledge, this is the first study to implement this clinical redefinition within a video-based system for the assessment of PD motor symptoms. The main contributions of this paper can be categorized into three key areas:  
	\begin{enumerate}
		\item Designing features corresponding to the four core components: bradykinesia, hypokinesia, sequence effect, and hesitation-halts.
		\item Redefining new sub-components through data-driven analysis, and
		\item Classifying MDS-UPDRS scores using machine learning.
	\end{enumerate}
	We elaborate on each of these findings in detail.\\
	We designed 13 features and analyzed their distribution across different levels of the MDS-UPDRS. Most features clearly distinguished between these severity levels. However, the amplitude sub-component of sequence effect did not exhibit a linear decline with increasing severity. As shown in Figure~\ref{fig:box_plots}, this feature followed a non-linear pattern— from UPDRS score 0 to 2, the interquartile is decreasing and from UPDRS 2 to 4 it is slightly increasing. A similar pattern was reported by Guarin et al.\cite{guarin2024trained}, where amplitude decreased up to UPDRS score 2 and then increased. We also examined the amplitude decrement features proposed by Guarin et al.\cite{guarin2024trained} and Islam et al.~\cite{islam2023using} on our dataset. Notably, all eight amplitude decrement features proposed by Islam et al.~\cite{islam2023using} and one feature proposed by Guarin et al.\cite{guarin2024trained} also exhibited a non-linear trend. These finding suggest that the amplitude sequence effect does not consistently worsen with disease severity, and patients at higher severity levels may not demonstrate greater amplitude decrement. \\
	To further explore the structure of motor impairments, we applied PCA with varimax rotation. The results aligned partially with the four proposed motor components but revealed additional substructures. In particular, hesitation-halts may reflect separate irregularities in amplitude, timing, and speed, while the sequence effect may involve distinct substructures in progressive amplitude decrement, progressive speed decrement, and progressive increment of tapping intervals. These findings suggest that a more granular categorization of motor deficits could improve both the sensitivity and clinical utility of automated assessments. Future research is needed to establish the clinical relevance of these components by evaluating their ability to capture treatment effects and track disease progression.\\
	Since clinician-based assessment through MDS-UPDRS is currently the gold standard for quantifying motor impairment in PD ~\cite{guerra2023objective}, a machine learning model is also trained on the extracted features to predict the MDS-UPDRS score. This may serve as a validation of the proposed features through their alignment with clinical scores. Our results outperformed existing video-based methods mostly due to the use of distance-based signal, which are more robust to camera viewpoint variability compared to angle-based signal commonly used in previous works ~\cite{guarin2024trained, islam2023using}. This finding suggests that distance-based signal in the tapping test is more reliable and accurate for motor impairment quantification. Finally, our SHAP analysis revealed that average tapping interval in the bradykinesia category, speed and amplitude variability in the hesitation-halts category were the most influential in determining the MDS-UPDRS score—highlighting which aspects of motor dysfunction are most emphasized in clinical assessments. \\
	Strengths of this work include its granularity, clinical alignment, less sensitivity to camera viewpoints, and improved MDS-UPDRS prediction performance. Importantly, our method does not require any physical contact such as sensor with the patient, making it a low-cost and convenient solution for both clinical and at-home use. Certain limitations must also be acknowledged. In some videos, finger keypoints may not be fully visible, which reduces the accuracy of the extracted features and affects the quality of the distance-based signal. To address this, we plan to implement a data quality module that alerts patients if their hands are not in an optimal position. Based on this feedback, patients can adjust their positioning to ensure accurate data collection. Additionally, for patients with severe impairment, the inability to perform the task properly may lead to inaccurate hand keypoint detection. For example, some patients could not raise their hands and place them on the chair’s armrest, causing keypoints to be partially occluded. In addition, although MDS-UPDRS scores were assigned by trained assessors, the subjectivity of clinical ratings introduces uncertainty in the ground truth. However, this is also the reason that developing machine learning models to predict MDS-UPDRS scores was not the main aim of our study, but rather to provide granular features of different motor characteristics. Lastly, while our dataset currently includes 485 videos, it represents only a subset of a larger longitudinal cohort. More video recordings will become available in the near future. Increasing the dataset size will help to improve robustness across varying recording conditions and enable further validation and  development of advanced models.	\\
	This framework offers versatile applications across clinical trials and patient care. In the context of clinical trials, our system can be used to objectively monitor motor impairments in patients with PD. This monitoring system may also be extended to individuals in the prodromal phase of PD. Detecting and tracking changes in early stage of PD is especially important for the clinical trials and tools that can sensitively capture small motor deficits will be essential for identifying treatment effects. In terms of patient care, the proposed framework supports telemedicine by enabling remote, standardized monitoring of motor symptom severity and response to treatment. This is particularly beneficial for patients with limited mobility, those living in rural or underserved areas, or individuals who face difficulties attending frequent in-clinic visits. In addition, proposed framework may aid in differential diagnosis, where distinguishing between conditions with overlapping motor symptoms may be challenging. For example, it could help differentiate PD from other movement disorders such as dystonic tremor. By providing objective, fine-grained measurements of motor patterns, the framework adds valuable clinical information that can support more accurate diagnosis and personalized care. \\
	Future research will focus on evaluating the responsiveness of extracted features to medication effects and their potential to track disease progression over time.\\
	\section*{Methodology}
	\label{sec:methodology}
    \subsection*{Data}
	Data was obtained from the Personalized Parkinson Project (PPP). Details about this study can be found in~\cite{bloem2019personalized}. In brief, PPP is a cohort study including 520 early-stage (diagnosed $<5$ years ago) PD patients. Follow-up included in-clinic visits at baseline, after one year, and after 2 years, and continuous at home monitoring using a wrist-worn sensor for two up to three years. In this work, we used the video-recorded MDS-UPDRS part III examinations obtained during the yearly in-clinic visits. 
    
    As part of the MDS-UPDRS part III examination, each participant completed a finger-tapping task twice: once in an "off" state and once in an "on" state. To ensure an accurate off state assessment, participants abstained from dopamine-based medications for at least 12 hours before the first recording. After completing the first recording, participants took their prescribed medication, allowing time for it to take effect, and then performed the second finger-tapping test in the on state. The video recordings were conducted using a camera with a resolution of 1280×720 pixels at 25 frames per second (fps). 
    
    The in-person clinical evaluation of the finger-tapping tasks was performed by trained assessors following the MDS-UPDRS part III instructions. In addition, a subset of video recordings was also evaluated by trained assessors based on the video recordings. The current dataset comprises 485 videos, obtained from 74 subjects, of which 241 videos are recorded in the off state and 244 videos are recorded in the on state. 7 videos were evaluated by two assessors, 29 by three assessors, 2 by five assessors, and 470 by a single assessor. To assess the consistency among different clinical raters, we computed Krippendorff’s alpha, a statistical measure suitable for evaluating inter-rater reliability when the number of raters differs. The resulting value of 0.62 indicates reasonable but not perfect agreement, reinforcing the need for automated and reproducible scoring tools to enhance consistency and reduce variability in clinical assessments.

    In cases of multiple scores, the median score was used. A total of 23 videos were excluded from the dataset: 4 due to a duration of less than 8 seconds and 19 because the participant’s hands were cut off during recording. Table~\ref{table:demographics} summarizes the demographic and clinical characteristics of the participants, including age, sex distribution, disease duration, and medication status. Additionally, it presents scores from the MDS-UPDRS parts I, III, and IV, which assess different aspects of PD progression. The distribution of MDS-UPDRS scores is also detailed in Table~\ref{table:score_distribution}.
    \begin{table}[ht]
		\small
		\caption{Demographics and clinical information of patients in our dataset. Part I evaluates non-motor characteristics, including cognitive impairment, mood disturbances, sleep problems, and autonomic dysfunction. Part III consists of a motor examination conducted by a clinician, assessing rigidity, tremor, bradykinesia, and postural instability. Part IV covers motor complications, including dyskinesias and symptom fluctuations related to medication use. The Hoehn \& Yahr scale~\cite{hoehn1967parkinsonism}, a widely used measure for categorizing disease progression, provides further insight into the severity levels within the dataset.}
		\label{table:demographics}
		\centering
		\resizebox{\columnwidth}{!}{%
			\begin{tabular}{p{6cm}p{2cm}} 
				\hline
				\textbf{characteristicss} & \textbf{Value} \\
				\hline
				Age (Mean ± SD) & 61 ± 8.4 \\
				Sex, Man/Woman & 50 / 24 \\
				Disease duration (Mean ± SD) & 34 ± 17 \\
				Medication status, On/Off & 241 / 244 \\
				MDS-UPDRS I (Mean ± SD) & 2.85 ± 2.36 \\
				MDS-UPDRS III pre-med (Mean ± SD) & 37.16 ± 12.76 \\
				MDS-UPDRS III post-med (Mean ± SD) & 33.33 ± 13.01 \\
				MDS-UPDRS IV (Mean ± SD) & 3.25 ± 3.19 \\
				Hoehn \& Yahr stage pre-med (Mean ± SD) & 2.05 ± 0.76 \\
				Hoehn \& Yahr stage post-med (Mean ± SD) & 1.82 ± 0.70 \\
				\hline
			\end{tabular}
		}
	\end{table}
	
	\begin{table}[ht]
		\centering
		\small
		\caption{Score Distribution of Clinical MDS-UPDRS Scores}
		\resizebox{\columnwidth}{!}{%
			\begin{tabular}{cccccc}
				\toprule
				Score & 0 & 1 & 2 & 3 & 4 \\ 
				\midrule
				Number of videos & 17 & 60 & 177 & 219 & 12 \\ 
				Percentage (\%) & 3.5\% & 12.3\% & 36.4\% & 45.1\% & 2.4\% \\ 
				\bottomrule
			\end{tabular}
		}
		\label{table:score_distribution}
	\end{table}

	\subsection*{Video Preprocessing}
	\label{sec:Video Acquisition and Preprocessing}
	In the preprocessing step, the videos are cropped to focus on the patient and remove redundant background information. The cropping procedure is outlined in Algorithm~\ref{alg:video_cropping}. We applied the YOLOv5~\cite{glenn2020yolov5} person detection model to the first frame of each video to detect individuals and assign unique IDs to each person. Subsequently, a person re-identification (ReID) technique~\cite{zhou2019omni} was used to match detected individuals across frames and track their positions over time. For each detected individual, the number of frames in which they were consistently present was calculated. The individual with the longest continuous presence was identified as the patient. To define the cropping region, we selected the bounding box corresponding to the 90th percentile of the patient’s detected bounding box sizes across all frames. This approach helps to minimize the impact of potential tracking errors by excluding outlier bounding boxes. This bounding box was then enlarged by 20\% and adjusted to a square shape. Finally, all frames were cropped according to this adjusted bounding box, resulting in patient-centered video sequences.  
	
	\begingroup
	\begin{algorithm}[h!]
		\small
		\caption{Video Cropping}\label{alg:video_cropping}
		\begin{algorithmic}[1]
			\STATE \textbf{Input:} Video file, CSV file with video names
			\STATE \textbf{Output:} Cropped video focused on the patient	
			\STATE Load YOLOv5 model for person detection
			\STATE Load re-identification model (ResNet50) for tracking		
			\STATE Load video names from CSV
			\FOR{each video in video names}
			\STATE Initialize video capture and set output paths
			\STATE Process first frame to detect persons and assign IDs
			\STATE Extract features for each detected person in first frame		
			\WHILE{video has frames}
			\STATE Detect persons in the current frame
			\STATE Extract features for each detected person
			\FOR{each initially detected person}
			\STATE Match current frame's detected persons with initial features
			\IF{person is detected and matches criteria}
			\STATE Update last valid position and save bounding box
			\ENDIF
			\ENDFOR
			\ENDWHILE
			\STATE Select the most frequently detected person ID as patient
			\STATE Select bounding box at 90th percentile of area size
			\STATE Increase bounding box size by 20\%
			\STATE Adjust bounding box to be square
			\STATE Center bounding box around the patient
			\STATE Reinitialize video capture for cropping
			\STATE Initialize output video with adjusted bounding box dimensions
			\WHILE{video has frames}
			\STATE Crop each frame to fixed bounding box
			\STATE Write cropped frame to output video
			\ENDWHILE
			\ENDFOR
			
		\end{algorithmic}
	\end{algorithm}
	\endgroup

	\subsection*{Extraction of keypoints} 
	\label{sec:PD Motor characteristics Quantification}
    In order to quantify different motor characteristics, we first utilized MediaPipe~\cite{lugaresi2019mediapipe} to detect 21 keypoints of the hand in each video frame (as illustrated in Figure~\ref{fig:signal_generation}). Then, the Euclidean distance between the \textit{thumb} and \textit{index finger tip} is computed as follows:\\
	
	\small 
	\begin{equation}
		d_1(f_i) = \sqrt{(\text{TF}_x(i) - \text{IF}_x(i))^2 + (\text{TF}_y(i) - \text{IF}_y(i))^2}
	\end{equation}
	\normalsize 
	
	Where \(({TF}_x(i), {TF}_y(i)) \) are the x and y coordinates of the thumb tip and  \(({IF}_x(i), {IF}_y(i)) \) the x and y coordinates of the index finger tip at frame  \(i\). The distance \( d_1(i) \) is computed for every frame \( i \) in the video. To make the metric invariant to the distance from the camera, it is scaled by the distance between the index finger's metacarpophalangeal (MCP) joint as shown in Figure~\ref{fig:angle_dis} and the wrist keypoint as follows:\\
	
	\small 
	\begin{equation}
		d_2(f_i) = \sqrt{(\text{W}_x(i) - \text{IMCP}_x(i))^2 + (\text{Wrist}_y(i) - \text{IMCP}_y(i))^2}
	\end{equation}
	\normalsize 
	
	Here  \(({W}_x(i), {W}_y(i)) \) and \(({IMCP}_x(i), {IMCP}_y(i)) \) describe the x and y coordinates of wrist and index MCP, respectively.
	The scaled distance for each frame \( i \) is then calculated as:\\
	\small 
	\begin{equation}
		s(i) = \frac{d_1(i)}{d_2(i)}
	\end{equation}
	\normalsize 
	The distance signal over the entire video is obtained by concatenating the scaled distances across all $N$ frames:\\
	\small 
	\begin{equation}
		S = [s(1), s(2), \dots, s(N)]
		\label{eq:distance_signal}
	\end{equation}
	\normalsize 

    \subsection*{PD Motor characteristics Quantification} 
	\label{sec:PD Motor characteristics Quantification}
    Based on the recent update in the clinical definition of bradykinesia, we defined 10 interpretable signal features to capture the four primary PD motor characteristics captured during the finger-tapping examination: hypokinesia, bradykinesia, sequence effect and hesitation-halts~\cite{bologna2023redefining}. In addition, we also derived two features focused on movement speed, which capture the combined effects of hypo- and bradykinesia. Below, we describe how these features are derived from the scaled signal \( S \).
	
	\subsubsection*{Hypokinesia quantification}
    
    Hypokinesia refers to a reduced amplitude of movement (difficulty to fully open the hand). To obtain this amplitude, peaks and troughs are detected within the distance signal of equation~\eqref{eq:distance_signal} and the difference between a trough and a peak is defined as the amplitude. A peak detection algorithm~\cite{virtanen2020scipy} is applied to the scaled distance signal to identify peaks \( \{p_1, p_2, \dots, p_M\} \), and trough values \( \{t_1, t_2, \dots, t_K\} \) where \( M \) and \( K \)are the total number of detected peaks and troughs, respectively. These peaks correspond to frames when the patient fully opens their hand during tapping, whereas the troughs correspond to moments when the hand is closed. The tapping amplitudes are then computed as follows:\\
	
	\small
	\begin{equation}
		\mathrm{Amp}_{\mathrm{i}} = \mathrm{p}_i - \mathrm{t}_i, \quad i = 1, 2, \dots, \min(M, K)
	\end{equation}
	\normalsize
	Based on this data, two key features are extracted to quantify hypokinesia:
	\small 
	\begin{itemize}
		\item Maximum Amplitude\\ It represents the maximum distance between the index and thumb fingers across all taps and it reflects the patient’s ability to open their hand widely:\\
	\end{itemize}	
	
	\small 
	\begin{equation}
		\mathrm{Amp}_{\mathrm{max}} = \max(\mathrm{Amp}_i), \quad i = 1, 2, \dots, N
	\end{equation}
	\normalsize 
	\begin{itemize}
		\item Average amplitude\\ It indicates the average amplitude across all taps, capturing the the overall amplitude of tapping.\\
	\end{itemize}	
	
	\small 
	\begin{equation}
		\mathrm{Amp}_{\mathrm{avg}} = \frac{1}{N} \sum_{i=1}^{N} \mathrm{Amp}_i
	\end{equation}
	\normalsize
    
	\subsubsection*{Bradykinesia quantification}
	Bradykinesia refers to the slowness of voluntary movements, which results in an increased tapping interval. We defined the tapping interval as the time between consecutive peaks in the distance signal. From this, we derived the average tapping interval:  
	
	\begin{itemize}
		\item Average Tapping Interval\\This feature measures the average time between successive peaks in the signal and it is calculated as follows:\\
	\end{itemize}	
	\small 
	\begin{equation}
		\mathrm{TI}_{\mathrm{avg}} = \frac{\sum_{j=2}^{M} \left( T_{p_j} - T_{p_{j-1}} \right)}{M - 1}
	\end{equation}
	\normalsize 
	
	where \( T_{p_j} \) is the timestamp of the \( j \)-th peak.
    
	\subsubsection*{Combined Hypokinesia-Bradykinesia Quantification}
	Because there is evidence that brady- and hypokinesia share a common pathophysiological background, and both improve with dopaminergic medication or deep brain stimulation, we added two combined brady- and hypokinesia features derived from the speed signal. Since speed is influenced by both the amplitude and tapping interval, it is a combination of hypokinesia and bradykinesia. 

    The speed signal is computed as the displacement between consecutive frames divided by the time between two frames and it quantifies how quickly the hands move over time: \\
	\small
	\begin{equation}
		\mathrm{speed}_i = \frac{\lvert \mathrm{s}_i - \mathrm{s}_{i-1} \rvert}{\Delta t},
		\quad
		\Delta t = \frac{1}{\mathrm{FPS}}
	\end{equation}
	\normalsize
	where \( FPS \) is video frame rate. 

    Two features are obtained in this category according to the following equations: 

    \begin{itemize}
		\item Cycle Average Speed (CAS)\\ The average speed for each tapping cycle is obtained as follows:\\
	\end{itemize}
	
	\small
	\begin{equation}
		\mathrm{CAS}_j
		= \frac{1}{\bigl(T_{p_j} - T_{p_{j-1}}\bigr)} 
		\sum_{t = T_{p_{j-1}} + \Delta t }^{T_{p_j}} \mathrm{Speed}(t),
		\quad j = 1, 2, \dots, M-1.
	\end{equation}
	\normalsize
	
	and the feature is obtained through averaging across all cycles as follows:\\
	
	\small 
	\begin{equation}
		\mathrm{CAS}_{\mathrm{avg}} = \frac{1}{M} \sum_{i=1}^{M-1} \mathrm{CAS}_i
	\end{equation}
	\normalsize 
    This feature captures the average rate at which the position of the fingers changes over time and depends on both the amplitude and the tapping interval. 
    \begin{itemize}
		\item Cycle Maximum Speed (CMS)\\ 	For each tapping cycle, 95\% speed values are obtained to account for incidental outliers that may occur due to keypoint detection:\\
	\end{itemize}

	\small
	\begin{equation}
		\begin{aligned}
			\mathrm{CMS}_j
			&= \operatorname{percentile}_{0.95}
			\Bigl(\{\mathrm{Speed}(t) : T_{p_{j-1}} \le t \le T_{p_j}\}\Bigr),\\
			&\quad j = 1, 2, \dots, M-1.
		\end{aligned}
	\end{equation}
	\normalsize

	The metric is averaged across all cycles:\\
	\small 
	\begin{equation}
		\mathrm{CMS}_{\mathrm{avg}} = \frac{1}{M} \sum_{i=1}^{M-1} \mathrm{CMS}_i
	\end{equation}
	\normalsize 
	This features captures the typical maximum speed achieved across tapping cycles. This feature is related to the amplitude and tapping interval, but also captures additional information because the maximum speed also depends on the shape of the tapping movement. 
    
	\subsubsection*{Sequence effect quantification}
    Sequence effect is defined as a progressive reduction in amplitude and/or speed, and progressive increase in tapping interval during the task. Patients with PD often struggle to maintain consistent performance during repetitive tasks such as tapping.
    We separately quantified the progressive reduction in amplitude and speed and the progressive increase in tapping intervals.
	\begin{itemize}
		\item Amplitude slope\\ A linear trend is fitted to the sequence of amplitudes \( amp_i \). The slope quantifies the rate at which the amplitude declines over time.
		\item Tapping interval slope\\ A linear trend is fitted to the sequence of tapping intervals. An increasing slope reflects that the patient takes progressively more time to complete each tap.
		\item Speed slope\\ A linear trend is fitted to the sequence of average speeds \( CAS_j \). The slope quantifies the rate at which average speed declines over time.
	\end{itemize}
	
	\subsubsection*{Hesitation-halts quantification}
	Hesitations-halts refer to increased variability in the regularity and timing of the tapping movements. We designed five features to separately capture the variability in amplitude, tapping interval, cycle average speed, cycle maximum speed, and the occurrence of longer interruptions:
	
	\begin{itemize}
		\item COV of Amplitude\\This feature captures irregularities in the tapping amplitude (normalized by the average amplitude), and is calculated as:\\
	\end{itemize}	
	\small 
	\begin{equation}
		\mathrm{Amp}_{\mathrm{COV}} = \frac{\sqrt{\frac{\sum_{j=1}^{M} \left( p_j - \mathrm{amp}_{\mathrm{Avg}} \right)^2}{M}}}{\mathrm{amp}_{\mathrm{Avg}}}
        \end{equation}
	
	\normalsize 
	
	\begin{itemize}
			\item COV of Tapping Intervals\\This feature reflects irregularities in tapping interval (normalized by the average tapping interval), and is calculated as:\\
	\end{itemize}	
	
	\small
	\begin{equation}
		\mathrm{TI}_{\mathrm{COV}} = \frac{\sqrt{\frac{\sum_{j=2}^{M} \left( T_{p_j} - T_{p_{j-1}} - \mathrm{TI}_{\mathrm{avg}} \right)^2}{M - 1}}}{\mathrm{TI}_{\mathrm{avg}}}
	\end{equation}
	\normalsize 
	
	\begin{itemize}
		\item COV of speed\\ The following features are proposed to capture irregularities in maximum and average speed across tapping cycles:\\
	\end{itemize}	
	
	\small
	\begin{equation}
		\mathrm{CMS}_{\mathrm{COV}} = \frac{\sqrt{\frac{\sum_{j=1}^{M-1} \left( CMS_{j} - \mathrm{CMS}_{\mathrm{avg}} \right)^2}{M-1}}}{\mathrm{CMS}_{\mathrm{avg}}}
	\end{equation}
	\normalsize 
	
	\small
	\begin{equation}
		\mathrm{CAS}_{\mathrm{COV}} = \frac{\sqrt{\frac{\sum_{j=1}^{M-1} \left( CAS_{j} - \mathrm{CAS}_{\mathrm{avg}} \right)^2}{M-1}}}{\mathrm{CAS}_{\mathrm{avg}}}
	\end{equation}
	\normalsize 
	\begin{itemize}
		\item Number of interruptions\\
		Patients with PD may exhibit brief pauses, halts, or freezing during repetitive movements such as tapping~\cite{goetz2008movement}. To quantify them, we consider a tapping interval as an interruption if it is larger than a threshold value. The threshold value is equal to the two times the median for all tapping intervals. Each interval exceeding this threshold indicates a deviation from the patient’s typical rhythmic tapping. The total number of such intervals is summed to yield the \emph{Number of interruptions}.
		
	\end{itemize}	
	In summary, these features provide a comprehensive set of metrics to objectively quantify distinct groups of motor characteristics in PD. 

    \subsection*{Principal Component Analysis}
    	To investigate the underlying structure of the extracted video-based features and assess whether they align with the clinically defined categories of motor characteristics~\cite{bologna2023redefining}, we applied PCA followed by orthogonal varimax rotation~\cite{kaiser1958varimax}. PCA is a statistical method that transforms a set of potentially correlated variables into a set of linearly uncorrelated components, ranked by the amount of variance they capture from the original data. This allows for identifying latent dimensions that explain the co-variation structure among features while minimizing redundancy. Varimax rotation was utilized to enhance the interpretability of component loadings by maximizing the variance of squared loadings within each component. This rotation simplifies the factor structure and ensures each principal component is dominantly associated with a subset of features, which is critical for clinical interpretability. This dimensionality reduction approach allows us to empirically assess whether features presumed to belong to the same motor sign category (e.g., amplitude and speed decrements under the sequence effect) indeed exhibit high covariance, or whether they reflect distinct constructs. In doing so, PCA serves as a data-driven tool to validate or refine the clinical categorization of PD motor characteristics captured during the finger-tapping test. 
	
	\subsection*{Machine Learning Methods for Score Classification}
	\label{sec:classification}
	
	To verify the clinical validity of our approach, we evaluate whether the feature set can be used to train machine learning classifiers to estimate the MDS-UPDRS finger-tapping score. We evaluated three classifiers, logistic regression~\cite{menard2001applied}, LightGBM~\cite{ke2017lightgbm}, and random forest~\cite{breiman2001random}. Logistic regression is a simple linear classifier that delivers (pseudo-) probabilities for each class by comparing the weighted sum of input features across all classes. The final prediction corresponds to the class with the highest probability. The model's parameters are trained using maximum likelihood estimation (MLE) by minimizing the differences between predicted probabilities and actual labels. LightGBM constructs an ensemble of decision trees using gradient boosting, where each tree is trained to minimize the residual errors of its predecessors. The random forest classifier, in contrast, builds multiple independent decision trees by randomly sampling subsets of data points and features. The final prediction is derived from an average of all tree outputs, which typically delivers a low-variance model. We selected these classifiers to facilitate performance comparison, as logistic regression was used by Guarin et al.~\cite{guarin2024trained} and LightGBM was employed in the study by Islam et al.~\cite{islam2023using}. Additionally, we included the random forest classifier as a widely used non-linear ensemble method.

\begin{figure}[ht!]
		\centering
		\includegraphics[width=0.5\textwidth]{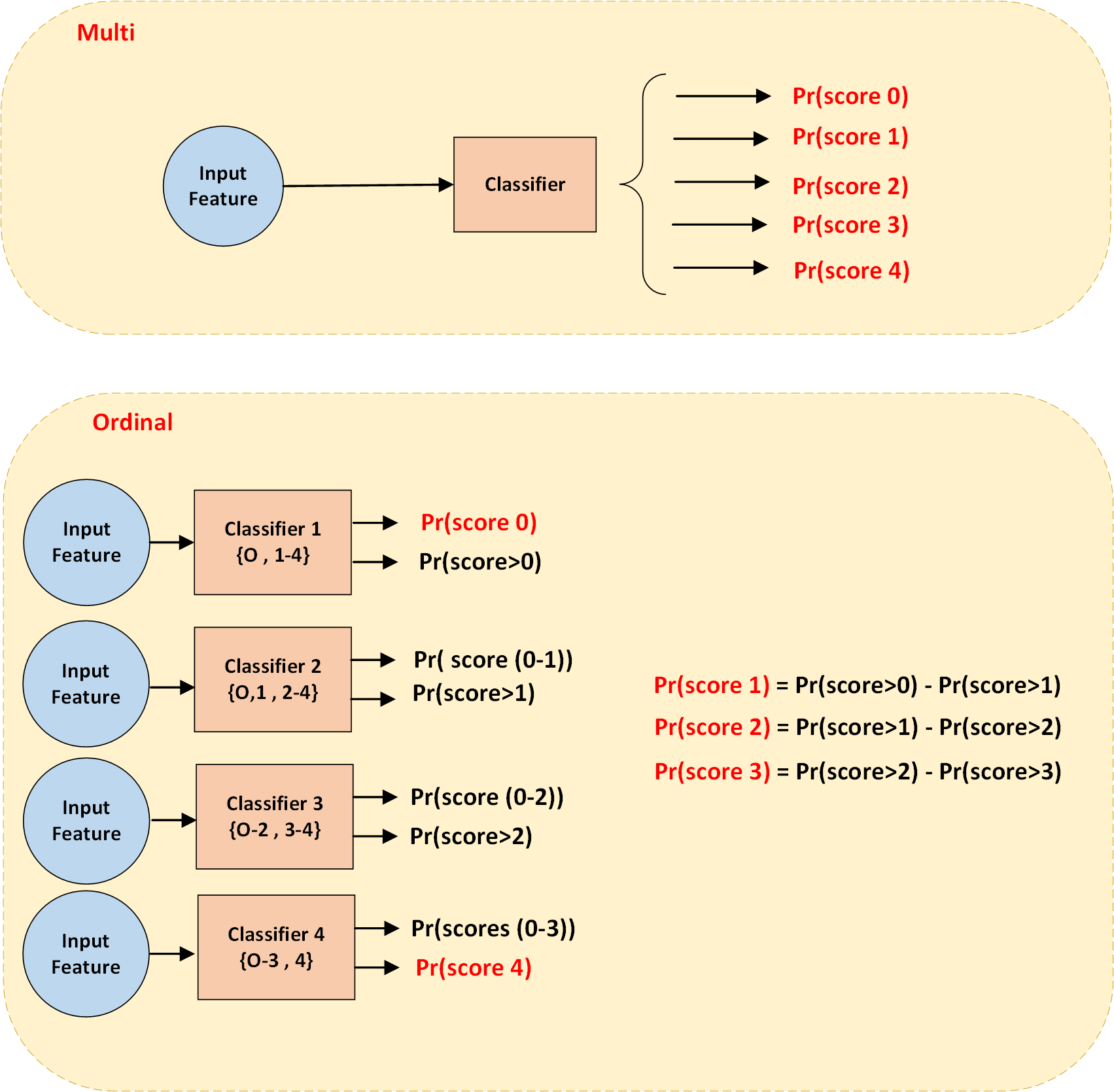} 
		\caption{\textbf{Multi-class and ordinal classification pipeline}. In multi-class classification, models directly predicted one of five MDS-UPDRS classes without considering any ordinal relationship. In ordinal classification, a sequence of four binary classifiers was trained: classifier 1 distinguishes class 0 from classes 1–4, classifier 2 distinguishes classes 0–1 from 2–4, classifier 3 distinguishes classes 0–2 from 3–4, and classifier 4 distinguishes classes 0–3 from 4.}
		\label{fig:multiordinalclassification}
	\end{figure}

	In this work, both multi-class classification and ordinal classification are employed to predict MDS-UPDRS scores~\cite{frank2001simple}. As shown in figure~\ref{fig:multiordinalclassification}, multi-class classification is used to predict one of several discrete and unordered classes. Each class is treated independently, with no assumptions about relationships or order between them. 
    On the other hand, ordinal classification considers the relative ordering between classes that might improve prediction consistency and accuracy by leveraging the ordinal structure. The approach is a decomposition of the ordinal classification into multiple binary classification problems. As shown in figure~\ref{fig:multiordinalclassification}, four binary classifiers are employed to model the ordinal relationships. The first classifier distinguishes between classes 1, 2, 3, 4 and class 0. The second classifier distinguishes between classes 2, 3, 4 and classes 0, 1. The third classifier distinguishes between classes 3, 4 and classes 0, 1, 2, 3. The final classifier distinguishes between classes 0, 1, 2, 3 and 4. The evaluation for ordinal classification involves combining their outputs to determine the final predicted class. 
	
\begin{figure}[ht!]
		\centering
		\includegraphics[width=0.4\textwidth]{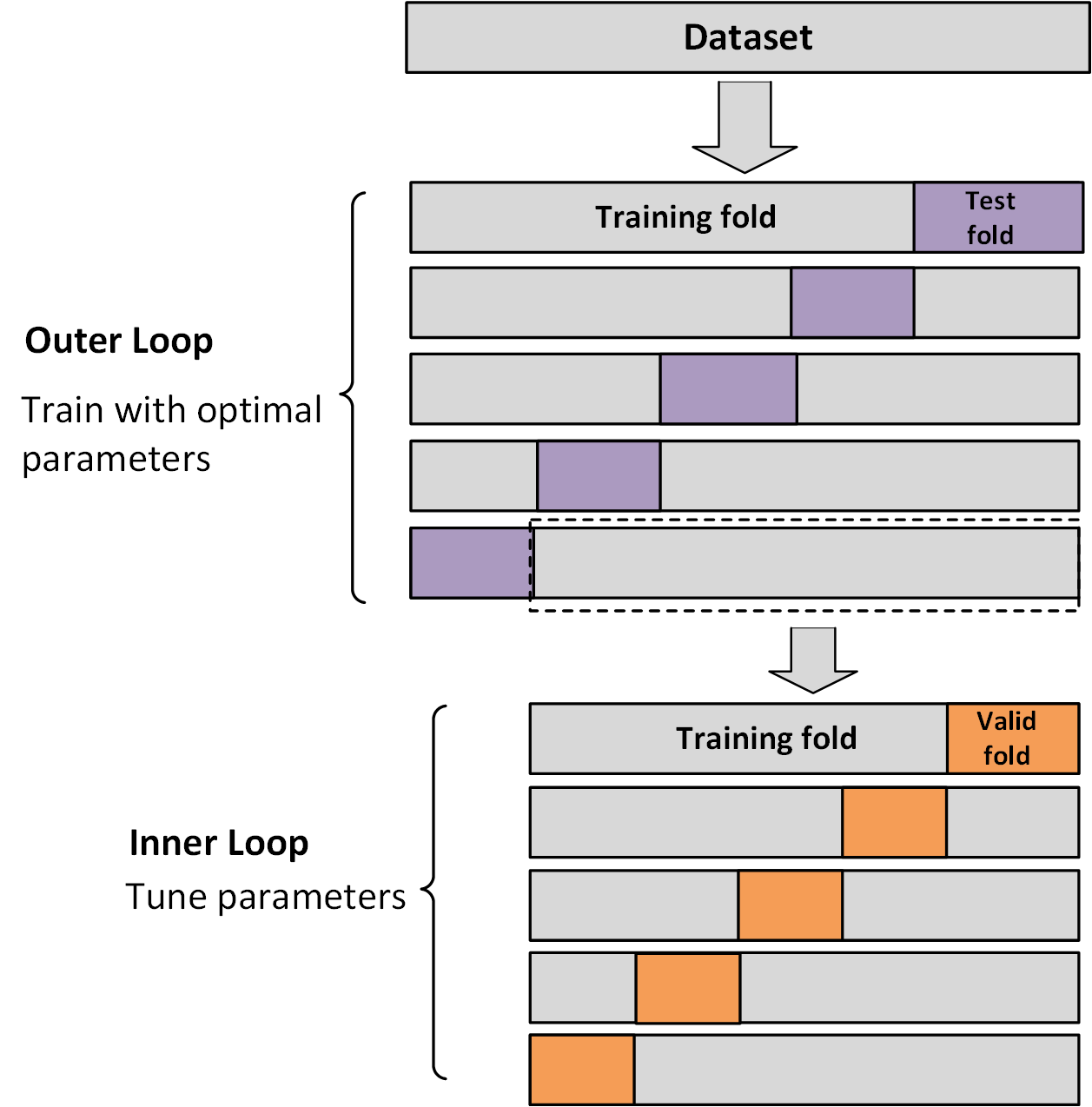} 
		\caption{\textbf{Nested cross-validation for hyperparameter optimization.} A 5-fold nested cross-validation scheme was used to optimize model hyperparameters. In each outer fold, data was split based on patient such that all videos from a given patient were assigned to either the training or the test set. The outer loop was used to evaluate model performance, while the inner loop performed cross-validation within the training set to select the best hyperparameters.}
		\label{fig:nested}
	\end{figure}

	For each classification algorithm, nested cross-validation, coupled with Optuna optimization~\cite{akiba2019optuna}, is utilized for hyperparameter tuning. As illustrated in Figure~\ref{fig:nested}, the process involves two levels of data splitting: an outer loop and an inner loop. In the outer loop, the dataset is divided into training and testing subsets using 5-fold cross-validation . Within the inner loop, the training subset is further partitioned into training and validation folds using 5-fold cross-validation for computational efficiency. Hyperparameter tuning is conducted within the inner folds, and the best-performing hyperparameters are applied for model evaluation in the outer loop. After completing the hyperparameter optimization process, five distinct hyperparameter sets are generated, one for each fold. Among these, the hyperparameter set achieving the highest accuracy is selected as the final configuration for subsequent LOOCV evaluation. All classifiers are implemented and trained using the Scikit-learn library~\cite{pedregosa2011scikit}.
	
	To thoroughly evaluate the effectiveness of our proposed MDS-UPDRS classification method, we conducted a comparative analysis with several state-of-the-art approaches from recent literature. For a fair and unbiased comparison, all related works are re-trained on our dataset, and their hyperparameters are optimized. This approach ensures that any observed performance differences arise from methodological variations rather than details in data or hyperparameter tuning. The first approach for comparison is selected from Islam et al.~\cite{islam2023using} with 116 features including speed, acceleration, amplitude, period, and frequency. A cross correlation between pair of features is applied to identify highly correlated pairs and reduce feature size. A multi-class lightGBM classifier is used to classify 5 classes of severity scores. The second approach for comparison is selected from Guarin et al.~\cite{guarin2024trained} with features from angular displacement signal including amplitude, speed, cycle duration, and movement rate. As classification in our dataset is 5-classes and tiered classification proposed by Guarin et al. is 4-classes, only multi-class and ordinal logistic regression classifications are considered for the comparison. We also compared our method with a deep learning-based approach by Lu et al.~\cite{lu2021quantifying} that uses raw distance and motion signals as input and trains a CNN-based model to automatically learn patterns in the data and classify five levels of severity. 

    \subsubsection*{Evaluation Metrics}
	\label{sec:metrics}
	As evaluation metrics we included accuracy, balanced accuracy, acceptable accuracy, macro precision, and macro F1-score. The final results were reported as the average performance across all folds. For consistency, the same hyperparameter settings were used across all folds during evaluation. The detailed metrics are defined as follows:
	
	1. Accuracy:  
	It measures the proportion of correctly predicted samples among the total number of samples: \\
	\small
	\[
	\text{Accuracy} = \frac{\sum_{i=1}^K TP_i}{\sum_{i=1}^K P_i}
	\]\\
	\normalsize
	where $TP_i$ is true positives for class $i$, representing the number of samples correctly predicted as belonging to class $i$.	 $P_i$ is total number of actual samples in class $i$ and $K$ is the total number of classes.
	
	Although accuracy is widely used, it is less suitable for highly imbalanced datasets. To address this limitation, \textit{balanced accuracy} is also included.
	
	2. Balanced Accuracy:  
	It accounts for class imbalance by averaging the accuracy for each class to provide a fair evaluation across all classes:\\
	\small
	\[
	\text{Balanced ACC} = \frac{1}{K} \sum_{i=1}^K \frac{TP_i}{P_i} = \frac{1}{K} \sum_{i=1}^K \frac{TP_i}{TP_i + FN_i}
	\]\\
	\normalsize
	where $FN_i$ is false negatives for class $i$, representing the number of samples from class $i$ incorrectly predicted as belonging to other classes. 
	
	3. Acceptable Accuracy: 
	This metric evaluates the proportion of predictions that fall within a tolerable range of the true labels, specifically predictions that are exactly one class below or above the true label:\\
	\small
	\[
	\text{Acceptable ACC} = \frac{N \left( |\hat{y} - y| \leq 1 \right)}{N} \times 100\%
	\]\\
	\normalsize
	where $\hat{y}$ is the predicted label and $y$ the true label for a sample, and $N$ denotes the total number of samples.
	This metric is particularly useful in scenarios where small deviations in predictions are still considered acceptable.
	
	4. Macro Precision:  
	Precision measures the proportion of true positive predictions among all positive predictions made by the model:\\
	\small
	\[
	\text{Precision}_i = \frac{TP_i}{TP_i + FP_i}
	\]\\
	\normalsize
	where $FP_i$ is false positives for class $i$, representing the number of samples incorrectly predicted as belonging to class $i$. The macro-averaged precision aggregates precision values across all classes to provide a single metric:\\
	\small
	\[
	\text{Precision}_{\text{Macro-average}} = \frac{\sum_{i=1}^K \text{Precision}_i}{K}
	\]
	\\
	\normalsize
	This metric is crucial in scenarios where minimizing false positives is important.
	
	5. Macro F1-Score:  
	The F1-score is the harmonic mean of precision and recall that offers a balanced evaluation of the two metrics. \\
	
	\small
	\[
	F1_i = \frac{\text{Precision}_i \times \text{Recall}_i}{\text{Precision}_i + \text{Recall}_i}
	\]
	
	\[
	F1_{\text{Macro-average}} = \frac{\sum_{i=1}^{K} F1_i}{K}
	\]	
     To express the uncertainty of our performance estimates, we report 95\% confidence intervals for each evaluation metric based on the variation across all cross-validation folds. The confidence intervals were calculated using the following formula: 
        
        \[
        CI = \bar{x} \pm 1.96 \cdot \frac{s}{\sqrt{n}}
        \]
        
        where $\bar{x}$ is the mean value of the metric across all folds, $s$ is the standard deviation across folds, and $n$ is the number of folds. The constant 1.96 corresponds to the critical value from the standard normal distribution for a 95\% confidence level.
        \subsubsection*{Statistical Significance Testing}
	To evaluate whether the observed differences in model performance between our approach and existing methods are statistically significant, we employed the Wilcoxon signed-rank test on each subject's individual performance ~\cite{wilcoxon1992individual}. This is a non-parametric paired difference test that assesses whether the median of the differences between paired observations is significantly different from zero. It is particularly suitable for our evaluation setting because:
	
	\begin{itemize}
		\item The performance metrics (e.g., accuracy, balanced accuracy) are obtained across multiple folds using LOOCV, which results in paired samples for each method.
		\item The data may not follow a normal distribution, which violates assumptions required by parametric tests such as the paired t-test.
	\end{itemize}
	
	A significance threshold of $\alpha = 0.05$ was used, and $p$-values below this threshold were considered statistically significant.

	\bibliographystyle{nature}
	\bibliography{references}

\begin{thebibliography}{10}
\expandafter\ifx\csname url\endcsname\relax
  \def\url#1{\texttt{#1}}\fi
\expandafter\ifx\csname urlprefix\endcsname\relax\def\urlprefix{URL }\fi
\providecommand{\bibinfo}[2]{#2}
\providecommand{\eprint}[2][]{\url{#2}}

\bibitem{feigin2019global}
\bibinfo{author}{Feigin, V.~L.} \emph{et~al.}
\newblock \bibinfo{title}{Global, regional, and national burden of neurological
  disorders, 1990--2016: a systematic analysis for the global burden of disease
  study 2016}.
\newblock \emph{\bibinfo{journal}{The Lancet Neurology}}
  \textbf{\bibinfo{volume}{18}}, \bibinfo{pages}{459--480}
  (\bibinfo{year}{2019}).

\bibitem{morris2000movement}
\bibinfo{author}{Morris, M.~E.}
\newblock \bibinfo{title}{Movement disorders in people with parkinson disease:
  a model for physical therapy}.
\newblock \emph{\bibinfo{journal}{Physical therapy}}
  \textbf{\bibinfo{volume}{80}}, \bibinfo{pages}{578--597}
  (\bibinfo{year}{2000}).

\bibitem{bologna2023redefining}
\bibinfo{author}{Bologna, M.} \emph{et~al.}
\newblock \bibinfo{title}{Redefining bradykinesia}.
\newblock \emph{\bibinfo{journal}{Movement Disorders}}
  \textbf{\bibinfo{volume}{38}}, \bibinfo{pages}{551} (\bibinfo{year}{2023}).

\bibitem{bologna2023further}
\bibinfo{author}{Bologna, M.} \& \bibinfo{author}{Guerra, A.}
\newblock \bibinfo{title}{Further insight into the role of primary motor cortex
  in bradykinesia pathophysiology} (\bibinfo{year}{2023}).

\bibitem{paparella2023may}
\bibinfo{author}{Paparella, G.} \emph{et~al.}
\newblock \bibinfo{title}{May bradykinesia features aid in distinguishing
  parkinson’s disease, essential tremor, and healthy elderly individuals?}
\newblock \emph{\bibinfo{journal}{Journal of Parkinson's disease}}
  \textbf{\bibinfo{volume}{13}}, \bibinfo{pages}{1047--1060}
  (\bibinfo{year}{2023}).

\bibitem{bologna2020there}
\bibinfo{author}{Bologna, M.} \emph{et~al.}
\newblock \bibinfo{title}{Is there evidence of bradykinesia in essential
  tremor?}
\newblock \emph{\bibinfo{journal}{European journal of neurology}}
  \textbf{\bibinfo{volume}{27}}, \bibinfo{pages}{1501--1509}
  (\bibinfo{year}{2020}).

\bibitem{passaretti2024role}
\bibinfo{author}{Passaretti, M.} \emph{et~al.}
\newblock \bibinfo{title}{The role of cerebellum and basal ganglia functional
  connectivity in altered voluntary movement execution in essential tremor}.
\newblock \emph{\bibinfo{journal}{The Cerebellum}}
  \textbf{\bibinfo{volume}{23}}, \bibinfo{pages}{2060--2081}
  (\bibinfo{year}{2024}).

\bibitem{ling2012hypokinesia}
\bibinfo{author}{Ling, H.}, \bibinfo{author}{Massey, L.~A.},
  \bibinfo{author}{Lees, A.~J.}, \bibinfo{author}{Brown, P.} \&
  \bibinfo{author}{Day, B.~L.}
\newblock \bibinfo{title}{Hypokinesia without decrement distinguishes
  progressive supranuclear palsy from parkinson's disease}.
\newblock \emph{\bibinfo{journal}{Brain}} \textbf{\bibinfo{volume}{135}},
  \bibinfo{pages}{1141--1153} (\bibinfo{year}{2012}).

\bibitem{laurencin2024noradrenergic}
\bibinfo{author}{Laurencin, C.} \emph{et~al.}
\newblock \bibinfo{title}{Noradrenergic alterations in parkinson’s disease: a
  combined 11c-yohimbine pet/neuromelanin mri study}.
\newblock \emph{\bibinfo{journal}{Brain}} \textbf{\bibinfo{volume}{147}},
  \bibinfo{pages}{1377--1388} (\bibinfo{year}{2024}).

\bibitem{goetz2008movement}
\bibinfo{author}{Goetz, C.~G.} \emph{et~al.}
\newblock \bibinfo{title}{Movement disorder society-sponsored revision of the
  unified parkinson's disease rating scale (mds-updrs): scale presentation and
  clinimetric testing results}.
\newblock \emph{\bibinfo{journal}{Movement disorders: official journal of the
  Movement Disorder Society}} \textbf{\bibinfo{volume}{23}},
  \bibinfo{pages}{2129--2170} (\bibinfo{year}{2008}).

\bibitem{richards1994interrater}
\bibinfo{author}{Richards, M.}, \bibinfo{author}{Marder, K.},
  \bibinfo{author}{Cote, L.} \& \bibinfo{author}{Mayeux, R.}
\newblock \bibinfo{title}{Interrater reliability of the unified parkinson's
  disease rating scale motor examination}.
\newblock \emph{\bibinfo{journal}{Movement Disorders}}
  \textbf{\bibinfo{volume}{9}}, \bibinfo{pages}{89--91} (\bibinfo{year}{1994}).

\bibitem{berlot2021variability}
\bibinfo{author}{Berlot, R.}, \bibinfo{author}{Rothwell, J.~C.},
  \bibinfo{author}{Bhatia, K.~P.} \& \bibinfo{author}{Kojovi{\'c}, M.}
\newblock \bibinfo{title}{Variability of movement disorders: the influence of
  sensation, action, cognition, and emotions}.
\newblock \emph{\bibinfo{journal}{Movement Disorders}}
  \textbf{\bibinfo{volume}{36}}, \bibinfo{pages}{581--593}
  (\bibinfo{year}{2021}).

\bibitem{angelini2024distinguishing}
\bibinfo{author}{Angelini, L.}, \bibinfo{author}{Paparella, G.} \&
  \bibinfo{author}{Bologna, M.}
\newblock \bibinfo{title}{Distinguishing essential tremor from parkinson’s
  disease: clinical and experimental tools}.
\newblock \emph{\bibinfo{journal}{Expert Review of Neurotherapeutics}}
  \textbf{\bibinfo{volume}{24}}, \bibinfo{pages}{799--814}
  (\bibinfo{year}{2024}).

\bibitem{guerra2023objective}
\bibinfo{author}{Guerra, A.}, \bibinfo{author}{D’Onofrio, V.},
  \bibinfo{author}{Ferreri, F.}, \bibinfo{author}{Bologna, M.} \&
  \bibinfo{author}{Antonini, A.}
\newblock \bibinfo{title}{Objective measurement versus clinician-based
  assessment for parkinson’s disease}.
\newblock \emph{\bibinfo{journal}{Expert Review of Neurotherapeutics}}
  \textbf{\bibinfo{volume}{23}}, \bibinfo{pages}{689--702}
  (\bibinfo{year}{2023}).

\bibitem{huckvale2019toward}
\bibinfo{author}{Huckvale, K.}, \bibinfo{author}{Venkatesh, S.} \&
  \bibinfo{author}{Christensen, H.}
\newblock \bibinfo{title}{Toward clinical digital phenotyping: a timely
  opportunity to consider purpose, quality, and safety}.
\newblock \emph{\bibinfo{journal}{NPJ digital medicine}}
  \textbf{\bibinfo{volume}{2}}, \bibinfo{pages}{1--11} (\bibinfo{year}{2019}).

\bibitem{tarolli2020feasibility}
\bibinfo{author}{Tarolli, C.~G.} \emph{et~al.}
\newblock \bibinfo{title}{Feasibility, reliability, and value of remote
  video-based trial visits in parkinson’s disease}.
\newblock \emph{\bibinfo{journal}{Journal of Parkinson’s disease}}
  \textbf{\bibinfo{volume}{10}}, \bibinfo{pages}{1779--1786}
  (\bibinfo{year}{2020}).

\bibitem{larson2021new}
\bibinfo{author}{Larson, D.~N.}, \bibinfo{author}{Schneider, R.~B.} \&
  \bibinfo{author}{Simuni, T.}
\newblock \bibinfo{title}{A new era: the growth of video-based visits for
  remote management of persons with parkinson’s disease}.
\newblock \emph{\bibinfo{journal}{Journal of Parkinson's disease}}
  \textbf{\bibinfo{volume}{11}}, \bibinfo{pages}{S27--S34}
  (\bibinfo{year}{2021}).

\bibitem{li2021automated}
\bibinfo{author}{Li, H.}, \bibinfo{author}{Shao, X.}, \bibinfo{author}{Zhang,
  C.} \& \bibinfo{author}{Qian, X.}
\newblock \bibinfo{title}{Automated assessment of parkinsonian finger-tapping
  tests through a vision-based fine-grained classification model}.
\newblock \emph{\bibinfo{journal}{Neurocomputing}}
  \textbf{\bibinfo{volume}{441}}, \bibinfo{pages}{260--271}
  (\bibinfo{year}{2021}).

\bibitem{williams2020supervised}
\bibinfo{author}{Williams, S.} \emph{et~al.}
\newblock \bibinfo{title}{Supervised classification of bradykinesia in
  parkinson’s disease from smartphone videos}.
\newblock \emph{\bibinfo{journal}{Artificial Intelligence in Medicine}}
  \textbf{\bibinfo{volume}{110}}, \bibinfo{pages}{101966}
  (\bibinfo{year}{2020}).

\bibitem{deng2024interpretable}
\bibinfo{author}{Deng, D.} \emph{et~al.}
\newblock \bibinfo{title}{Interpretable video-based tracking and quantification
  of parkinsonism clinical motor states}.
\newblock \emph{\bibinfo{journal}{npj Parkinson's Disease}}
  \textbf{\bibinfo{volume}{10}}, \bibinfo{pages}{122} (\bibinfo{year}{2024}).

\bibitem{yu2023clinically}
\bibinfo{author}{Yu, T.}, \bibinfo{author}{Park, K.~W.},
  \bibinfo{author}{McKeown, M.~J.} \& \bibinfo{author}{Wang, Z.~J.}
\newblock \bibinfo{title}{Clinically informed automated assessment of finger
  tapping videos in parkinson’s disease}.
\newblock \emph{\bibinfo{journal}{Sensors}} \textbf{\bibinfo{volume}{23}},
  \bibinfo{pages}{9149} (\bibinfo{year}{2023}).

\bibitem{khan2014computer}
\bibinfo{author}{Khan, T.}, \bibinfo{author}{Nyholm, D.},
  \bibinfo{author}{Westin, J.} \& \bibinfo{author}{Dougherty, M.}
\newblock \bibinfo{title}{A computer vision framework for finger-tapping
  evaluation in parkinson's disease}.
\newblock \emph{\bibinfo{journal}{Artificial intelligence in medicine}}
  \textbf{\bibinfo{volume}{60}}, \bibinfo{pages}{27--40}
  (\bibinfo{year}{2014}).

\bibitem{skaramagkas2023multi}
\bibinfo{author}{Skaramagkas, V.}, \bibinfo{author}{Pentari, A.},
  \bibinfo{author}{Kefalopoulou, Z.} \& \bibinfo{author}{Tsiknakis, M.}
\newblock \bibinfo{title}{Multi-modal deep learning diagnosis of parkinson’s
  disease—a systematic review}.
\newblock \emph{\bibinfo{journal}{IEEE Transactions on Neural Systems and
  Rehabilitation Engineering}} \textbf{\bibinfo{volume}{31}},
  \bibinfo{pages}{2399--2423} (\bibinfo{year}{2023}).

\bibitem{amo2024computer}
\bibinfo{author}{Amo-Salas, J.} \emph{et~al.}
\newblock \bibinfo{title}{Computer vision for parkinson’s disease evaluation:
  A survey on finger tapping}.
\newblock In \emph{\bibinfo{booktitle}{Healthcare}}, vol.~\bibinfo{volume}{12},
  \bibinfo{pages}{439} (\bibinfo{organization}{MDPI}, \bibinfo{year}{2024}).

\bibitem{zhao2020time}
\bibinfo{author}{Zhao, Z.} \emph{et~al.}
\newblock \bibinfo{title}{Time series clustering to examine presence of
  decrement in parkinson’s finger-tapping bradykinesia}.
\newblock In \emph{\bibinfo{booktitle}{2020 42nd Annual International
  Conference of the IEEE Engineering in Medicine \& Biology Society (EMBC)}},
  \bibinfo{pages}{780--783} (\bibinfo{organization}{IEEE},
  \bibinfo{year}{2020}).

\bibitem{heye2024validation}
\bibinfo{author}{Heye, K.} \emph{et~al.}
\newblock \bibinfo{title}{Validation of computer vision technology for
  analyzing bradykinesia in outpatient clinic videos of people with parkinson's
  disease}.
\newblock \emph{\bibinfo{journal}{Journal of the Neurological Sciences}}
  \textbf{\bibinfo{volume}{466}}, \bibinfo{pages}{123271}
  (\bibinfo{year}{2024}).

\bibitem{islam2023using}
\bibinfo{author}{Islam, M.~S.} \emph{et~al.}
\newblock \bibinfo{title}{Using ai to measure parkinson’s disease severity at
  home}.
\newblock \emph{\bibinfo{journal}{npj Digital Medicine}}
  \textbf{\bibinfo{volume}{6}}, \bibinfo{pages}{156} (\bibinfo{year}{2023}).

\bibitem{lugaresi2019mediapipe}
\bibinfo{author}{Lugaresi, C.} \emph{et~al.}
\newblock \bibinfo{title}{Mediapipe: A framework for building perception
  pipelines} (\bibinfo{year}{2019}).

\bibitem{guarin2024characterizing}
\bibinfo{author}{Guarin, D.~L.}, \bibinfo{author}{Wong, J.~K.},
  \bibinfo{author}{McFarland, N.~R.} \& \bibinfo{author}{Ramirez-Zamora, A.}
\newblock \bibinfo{title}{Characterizing disease progression in parkinson’s
  disease from videos of the finger tapping test}.
\newblock \emph{\bibinfo{journal}{IEEE Transactions on Neural Systems and
  Rehabilitation Engineering}}  (\bibinfo{year}{2024}).

\bibitem{guo2022vision}
\bibinfo{author}{Guo, Z.} \emph{et~al.}
\newblock \bibinfo{title}{Vision-based finger tapping test in patients with
  parkinson’s disease via spatial-temporal 3d hand pose estimation}.
\newblock \emph{\bibinfo{journal}{IEEE Journal of Biomedical and Health
  Informatics}} \textbf{\bibinfo{volume}{26}}, \bibinfo{pages}{3848--3859}
  (\bibinfo{year}{2022}).

\bibitem{redmon2018yolov3}
\bibinfo{author}{Redmon, J.}
\newblock \bibinfo{title}{Yolov3: An incremental improvement}.
\newblock \emph{\bibinfo{journal}{arXiv preprint arXiv:1804.02767}}
  (\bibinfo{year}{2018}).

\bibitem{lu2021quantifying}
\bibinfo{author}{Lu, M.} \emph{et~al.}
\newblock \bibinfo{title}{Quantifying parkinson’s disease motor severity
  under uncertainty using mds-updrs videos}.
\newblock \emph{\bibinfo{journal}{Medical image analysis}}
  \textbf{\bibinfo{volume}{73}}, \bibinfo{pages}{102179}
  (\bibinfo{year}{2021}).

\bibitem{yang2024fasteval}
\bibinfo{author}{Yang, Y.-Y.} \emph{et~al.}
\newblock \bibinfo{title}{Fasteval parkinsonism: an instant deep
  learning--assisted video-based online system for parkinsonian motor symptom
  evaluation}.
\newblock \emph{\bibinfo{journal}{npj Digital Medicine}}
  \textbf{\bibinfo{volume}{7}}, \bibinfo{pages}{31} (\bibinfo{year}{2024}).

\bibitem{yang2022automatic}
\bibinfo{author}{Yang, N.} \emph{et~al.}
\newblock \bibinfo{title}{Automatic detection pipeline for accessing the motor
  severity of parkinson’s disease in finger tapping and postural stability}.
\newblock \emph{\bibinfo{journal}{IEEE Access}} \textbf{\bibinfo{volume}{10}},
  \bibinfo{pages}{66961--66973} (\bibinfo{year}{2022}).

\bibitem{kaiser1958varimax}
\bibinfo{author}{Kaiser, H.~F.}
\newblock \bibinfo{title}{The varimax criterion for analytic rotation in factor
  analysis}.
\newblock \emph{\bibinfo{journal}{Psychometrika}}
  \textbf{\bibinfo{volume}{23}}, \bibinfo{pages}{187--200}
  (\bibinfo{year}{1958}).

\bibitem{guarin2024trained}
\bibinfo{author}{Guarín, D.~L.} \emph{et~al.}
\newblock \bibinfo{title}{What the trained eye cannot see: Quantitative
  kinematics and machine learning detect movement deficits in early-stage
  parkinson's disease from videos}.
\newblock \emph{\bibinfo{journal}{Parkinsonism \& Related Disorders}}
  \textbf{\bibinfo{volume}{127}}, \bibinfo{pages}{107104}
  (\bibinfo{year}{2024}).

\bibitem{bloem2019personalized}
\bibinfo{author}{Bloem, B.} \emph{et~al.}
\newblock \bibinfo{title}{The personalized parkinson project: examining disease
  progression through broad biomarkers in early parkinson’s disease}.
\newblock \emph{\bibinfo{journal}{BMC neurology}}
  \textbf{\bibinfo{volume}{19}}, \bibinfo{pages}{1--10} (\bibinfo{year}{2019}).

\bibitem{hoehn1967parkinsonism}
\bibinfo{author}{Hoehn, M.~M.} \& \bibinfo{author}{Yahr, M.~D.}
\newblock \bibinfo{title}{Parkinsonism: onset, progression, and mortality}.
\newblock \emph{\bibinfo{journal}{Neurology}} \textbf{\bibinfo{volume}{17}},
  \bibinfo{pages}{427--427} (\bibinfo{year}{1967}).

\bibitem{glenn2020yolov5}
\bibinfo{author}{Jocher, G.}
\newblock \bibinfo{title}{{YOLOv5} by ultralytics} (\bibinfo{year}{2020}).

\bibitem{zhou2019omni}
\bibinfo{author}{Zhou, K.}, \bibinfo{author}{Yang, Y.},
  \bibinfo{author}{Cavallaro, A.} \& \bibinfo{author}{Xiang, T.}
\newblock \bibinfo{title}{Omni-scale feature learning for person
  re-identification}.
\newblock In \emph{\bibinfo{booktitle}{Proceedings of the IEEE/CVF
  international conference on computer vision}}, \bibinfo{pages}{3702--3712}
  (\bibinfo{year}{2019}).

\bibitem{virtanen2020scipy}
\bibinfo{author}{Virtanen, P.} \emph{et~al.}
\newblock \bibinfo{title}{Scipy 1.0: fundamental algorithms for scientific
  computing in python}.
\newblock \emph{\bibinfo{journal}{Nature methods}}
  \textbf{\bibinfo{volume}{17}}, \bibinfo{pages}{261--272}
  (\bibinfo{year}{2020}).

\bibitem{menard2001applied}
\bibinfo{author}{Menard, S.}
\newblock \emph{\bibinfo{title}{Applied logistic regression analysis}}
  (\bibinfo{publisher}{SAGE publications}, \bibinfo{year}{2001}).

\bibitem{ke2017lightgbm}
\bibinfo{author}{Ke, G.} \emph{et~al.}
\newblock \bibinfo{title}{Lightgbm: A highly efficient gradient boosting
  decision tree}.
\newblock \emph{\bibinfo{journal}{Advances in neural information processing
  systems}} \textbf{\bibinfo{volume}{30}} (\bibinfo{year}{2017}).

\bibitem{breiman2001random}
\bibinfo{author}{Breiman, L.}
\newblock \bibinfo{title}{Random forests}.
\newblock \emph{\bibinfo{journal}{Machine learning}}
  \textbf{\bibinfo{volume}{45}}, \bibinfo{pages}{5--32} (\bibinfo{year}{2001}).

\bibitem{frank2001simple}
\bibinfo{author}{Frank, E.} \& \bibinfo{author}{Hall, M.}
\newblock \bibinfo{title}{A simple approach to ordinal classification}.
\newblock In \emph{\bibinfo{booktitle}{Machine Learning: ECML 2001: 12th
  European Conference on Machine Learning Freiburg, Germany, September 5--7,
  2001 Proceedings 12}}, \bibinfo{pages}{145--156}
  (\bibinfo{organization}{Springer}, \bibinfo{year}{2001}).

\bibitem{akiba2019optuna}
\bibinfo{author}{Akiba, T.} \emph{et~al.}
\newblock \bibinfo{title}{Optuna: A next-generation hyperparameter optimization
  framework}.
\newblock In \emph{\bibinfo{booktitle}{Proceedings of the 25th ACM SIGKDD
  International Conference on Knowledge Discovery \& Data Mining}},
  \bibinfo{pages}{2623--2631} (\bibinfo{year}{2019}).

\bibitem{pedregosa2011scikit}
\bibinfo{author}{Pedregosa, F.} \emph{et~al.}
\newblock \bibinfo{title}{Scikit-learn: Machine learning in python}.
\newblock \emph{\bibinfo{journal}{the Journal of machine Learning research}}
  \textbf{\bibinfo{volume}{12}}, \bibinfo{pages}{2825--2830}
  (\bibinfo{year}{2011}).

\bibitem{wilcoxon1992individual}
\bibinfo{author}{Wilcoxon, F.}
\newblock \bibinfo{title}{Individual comparisons by ranking methods}.
\newblock In \emph{\bibinfo{booktitle}{Breakthroughs in statistics: Methodology
  and distribution}}, \bibinfo{pages}{196--202} (\bibinfo{publisher}{Springer},
  \bibinfo{year}{1992}).

\end{thebibliography}
	\subsection*{ACKNOWLEDGEMENTS}
        We thank Debbie de Graaf for her help with re-scoring misclassified videos. This work was financially supported by the Dutch Research Council Long-Term Program (project \#KICH3.LTP.20.006, financed by the Dutch Research Council, Verily, and the Dutch Ministry of Economic Affairs and Climate Policy) and by a Donders research stimulation fund. Michael Tangermann received support from the DBI2 project (024.005.022, Gravitation), which is financed by the Dutch Ministry of Education (OCW) via the Dutch Research Council (NWO), from the Dareplane collaboration project which is co-funded by PPP Allowance awarded by Health Holland, Top Sector Life Sciences \& Health, and by a contribution from the Dutch Brain Foundation.
        The Center of Expertise for Parkinson and Movement Disorders was supported by a Center of Excellence grant from the Parkinson's Foundation.
	\subsection*{AUTHOR CONTRIBUTIONS}
        \textbf{Tahereh Zarrat Ehsan}: Writing – original draft, Data curation, Methodology, Software, Visualization. \textbf{Michael Tangermann}: Conceptualization, Methodology, Project administration, Supervision, Writing – review \& editing. \textbf{Yağmur Güçlütürk}: Conceptualization, Methodology, Project administration, Supervision, Writing – review \& editing. \textbf{Bastiaan R. Bloem}: Conceptualization, Methodology, Project administration, Supervision, Writing – review \& editing. \textbf{Luc J. W. Evers}: Conceptualization, Methodology, Project administration, Supervision, Writing – review \& editing.
        \subsection*{COMPETING INTERESTS}
        \textbf{Tahereh Zarrat Ehsan}, \textbf{Michael Tangermann},\textbf{Yağmur Güçlütürk}, and \textbf{Luc J. W. Evers} declare no competing interests. \textbf{Bastiaan R. Bloem} serves as the co-Editor in Chief for the Journal of Parkinson’s disease, serves on the editorial board of Practical Neurology and Digital Biomarkers, has received fees from serving on the scientific advisory board for the Critical Path Institute, Gyenno Science, MedRhythms, UCB, Kyowa Kirin and Zambon (paid to the Institute), has received fees for speaking at conferences from AbbVie, Bial, Biogen, GE Healthcare, Oruen, Roche, UCB and Zambon (paid to the Institute), and has received research support from Biogen, Cure Parkinson’s, Davis Phinney Foundation, Edmond J. Safra Foundation, Fred Foundation, Gatsby Foundation, Hersenstichting Nederland, Horizon 2020, IRLAB Therapeutics, Maag Lever Darm Stichting, Michael J Fox Foundation, Ministry of Agriculture, Ministry of Economic Affairs \& Climate Policy, Ministry of Health, Welfare and Sport, Netherlands Organization for Scientific Research (ZonMw), Not Impossible, Parkinson Vereniging, Parkinson’s Foundation, Parkinson’s UK, Stichting Alkemade-Keuls, Stichting Parkinson NL, Stichting Woelse Waard, Health Holland / Topsector Life Sciences and Health, UCB, Verily Life Sciences, Roche and Zambon.
        Prof. Bloem does not hold any stocks or stock options with any companies that are connected to Parkinson’s disease or to any of his clinical or research activities. 
        \subsection*{ADDITIONAL INFORMATION}
        \textbf{Correspondence} and requests for materials should be addressed to Tahereh Zarrat Ehsan.\\
    
\end{document}